\documentclass[lettersize,journal]{IEEEtran}

\usepackage{amssymb}
\usepackage{colortbl}
\usepackage{xcolor}
\usepackage{bm}
\usepackage{amsmath}
\usepackage{subfigure}
\usepackage{varwidth}
\usepackage{multirow}
\usepackage{graphicx}
\usepackage{algorithm}
\usepackage{algorithmic}
\usepackage{verbatim}
\usepackage{upgreek}
\usepackage{CJKutf8}
\usepackage{caption}
\usepackage{graphicx}
\usepackage{bmpsize}

\usepackage[caption=false,font=normalsize,labelfont=sf,textfont=sf]{subfig}

\DeclareGraphicsExtensions{.eps,.ps,.jpg,.bmp}

\begin{document}
\bibliographystyle{abbrv}

\title{{Illumination Controllable Dehazing Network based on Unsupervised Retinex Embedding}

\author{Jie Gui, Xiaofeng Cong, Lei He, Yuan Yan Tang, James Tin-Yau Kwok}

\thanks{J. Gui is with the School of Cyber Science and Engineering, Southeast University and with Purple Mountain Laboratories, Nanjing 210000, China (e-mail: guijie@seu.edu.cn).

X. Cong is with the School of Cyber Science and Engineering, Southeast University,  Nanjing 210000, China (e-mail: cxf\_svip@163.com).

L. He is with the Purple Mountain Laboratories,  Nanjing 210000, China (e-mail: helei@pmlabs.com.cn).

Y. Tang is with the Department of Computer and Information Science, University of Macau, Macau 999078, China (e-mail: yuanyant@gmail.com).

J. Kwok is with the Department of Computer Science and Engineering, The Hong Kong University of Science and Technology, Hong Kong 999077, China. (e-mail: jamesk@cse.ust.hk).
}}

\markboth{Journal of \LaTeX\ Class Files,~Vol.~14, No.~8, August~2021}%
{Shell \MakeLowercase{\textit{et al.}}: A Sample Article Using IEEEtran.cls for IEEE Journals}
\maketitle

\begin{abstract}
  On the one hand, the dehazing task is an ill-posedness problem, which means that no unique solution exists. On the other hand, the dehazing task should take into account the subjective factor, which is to give the user selectable dehazed images rather than a single result. Therefore, this paper proposes a multi-output dehazing network by introducing illumination controllable ability, called IC-Dehazing. The proposed IC-Dehazing can change the illumination intensity by adjusting the factor of the illumination controllable module, which is realized based on the interpretable Retinex theory. Moreover, the backbone dehazing network of IC-Dehazing consists of a Transformer with double decoders for high-quality image restoration. Further, the prior-based loss function and unsupervised training strategy enable IC-Dehazing to complete the parameter learning process without the need for paired data. To demonstrate the effectiveness of the proposed IC-Dehazing, quantitative and qualitative experiments are conducted on image dehazing, semantic segmentation, and object detection tasks. Code
  is available at {https://github.com/Xiaofeng-life/ICDehazing}.
\end{abstract}

\begin{IEEEkeywords}
  Image Dehazing, Illumination Controllable, Retinex, Unsupervised Prior, Transformer
\end{IEEEkeywords}

\IEEEpeerreviewmaketitle

\section{Introduction}
\begin{figure*}
  \centering
  \includegraphics[width=2.4cm,height=1.7cm]{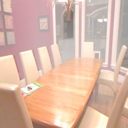}
  \includegraphics[width=2.4cm,height=1.7cm]{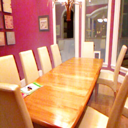}
  \includegraphics[width=2.4cm,height=1.7cm]{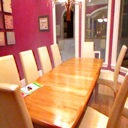}
  \includegraphics[width=2.4cm,height=1.7cm]{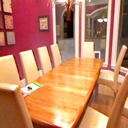}
  \includegraphics[width=2.4cm,height=1.7cm]{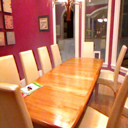}
  \includegraphics[width=2.4cm,height=1.7cm]{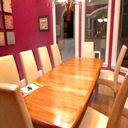}
  \includegraphics[width=2.4cm,height=1.7cm]{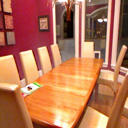}
  \vspace{0.1cm}
  \leftline{\hspace{0.8cm} (a) Hazy \hspace{1.2cm}  (b) 0.42 \hspace{1.0cm}   (c) 0.44  \hspace{1.2cm}  (d) 0.45  \hspace{1.2cm} (e) 0.46  \hspace{1.2cm}  (f) 0.47  \hspace{1.0cm} (g) 0.48}
  \\
  \includegraphics[width=2.4cm,height=1.7cm]{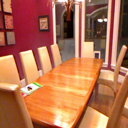}
  \includegraphics[width=2.4cm,height=1.7cm]{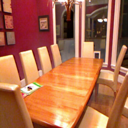}
  \includegraphics[width=2.4cm,height=1.7cm]{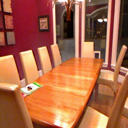}
  \includegraphics[width=2.4cm,height=1.7cm]{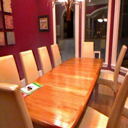}
  \includegraphics[width=2.4cm,height=1.7cm]{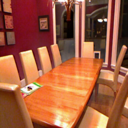}
  \includegraphics[width=2.4cm,height=1.7cm]{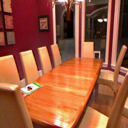}
  \includegraphics[width=2.4cm,height=1.7cm]{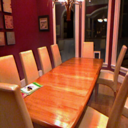}
  \\
  \leftline{\hspace{1.0cm}  (h) 0.49 \hspace{1.0cm}   (i) 0.50 \hspace{1.2cm}  (j) 0.51 \hspace{1.2cm} (k) 0.52 \hspace{1.2cm}  (l) 0.53  \hspace{1.2cm} (m) 0.54 \hspace{1.0cm} (n) 0.56}
  \caption{Visual dehazing results with different controllable parameters on SOTS indoor~\cite{li2018benchmarking}.}
  \label{fig:diff_alpha_ITS}
\end{figure*}

\begin{figure*}
  \centering
  \includegraphics[width=2.4cm,height=1.7cm]{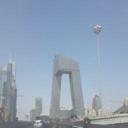}
  \includegraphics[width=2.4cm,height=1.7cm]{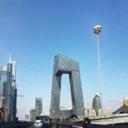}
  \includegraphics[width=2.4cm,height=1.7cm]{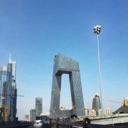}
  \includegraphics[width=2.4cm,height=1.7cm]{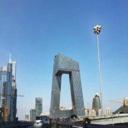}
  \includegraphics[width=2.4cm,height=1.7cm]{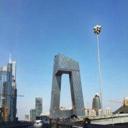}
  \includegraphics[width=2.4cm,height=1.7cm]{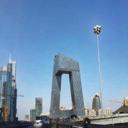}
  \includegraphics[width=2.4cm,height=1.7cm]{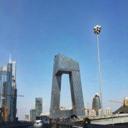}
  \vspace{0.1cm}
  \leftline{\hspace{0.8cm} (a) Hazy \hspace{1.2cm}  (b) 0.42 \hspace{1.0cm}   (c) 0.44  \hspace{1.2cm}  (d) 0.45  \hspace{1.2cm} (e) 0.46  \hspace{1.2cm}  (f) 0.47  \hspace{1.0cm} (g) 0.48}
  \\
  \includegraphics[width=2.4cm,height=1.7cm]{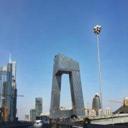}
  \includegraphics[width=2.4cm,height=1.7cm]{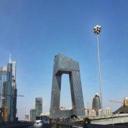}
  \includegraphics[width=2.4cm,height=1.7cm]{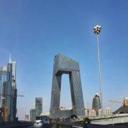}
  \includegraphics[width=2.4cm,height=1.7cm]{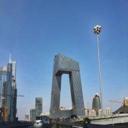}
  \includegraphics[width=2.4cm,height=1.7cm]{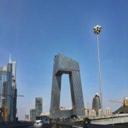}
  \includegraphics[width=2.4cm,height=1.7cm]{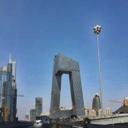}
  \includegraphics[width=2.4cm,height=1.7cm]{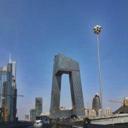}
  \\
  \leftline{\hspace{1.0cm}  (h) 0.49 \hspace{1.0cm}   (i) 0.50 \hspace{1.2cm}  (j) 0.51 \hspace{1.2cm} (k) 0.52 \hspace{1.2cm}  (l) 0.53  \hspace{1.2cm} (m) 0.54 \hspace{1.0cm} (n) 0.56}
  \caption{Visual dehazing results with different controllable parameters on SOTS outdoor~\cite{li2018benchmarking}.}
  \label{fig:diff_alpha_OTS}
\end{figure*}

\begin{figure*}
  \centering
  \includegraphics[width=2.4cm,height=1.7cm]{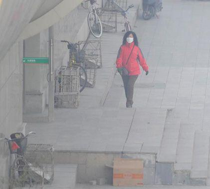}
  \includegraphics[width=2.4cm,height=1.7cm]{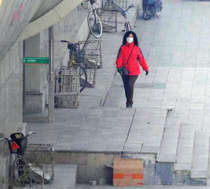}
  \includegraphics[width=2.4cm,height=1.7cm]{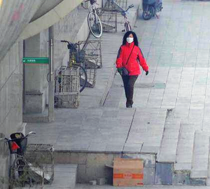}
  \includegraphics[width=2.4cm,height=1.7cm]{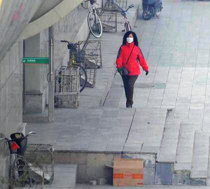}
  \includegraphics[width=2.4cm,height=1.7cm]{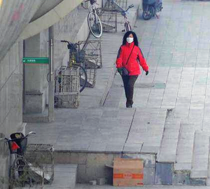}
  \includegraphics[width=2.4cm,height=1.7cm]{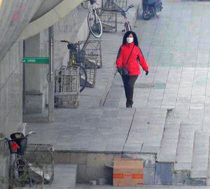}
  \includegraphics[width=2.4cm,height=1.7cm]{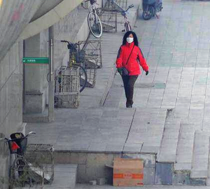}
  \vspace{0.1cm}
  \leftline{\hspace{0.8cm} (a) Hazy \hspace{1.2cm}  (b) 0.42 \hspace{1.0cm}   (c) 0.44  \hspace{1.2cm}  (d) 0.45  \hspace{1.2cm} (e) 0.46  \hspace{1.2cm}  (f) 0.47  \hspace{1.0cm} (g) 0.48}
  \\
  \includegraphics[width=2.4cm,height=1.7cm]{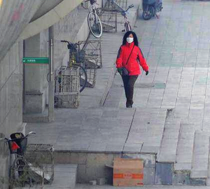}
  \includegraphics[width=2.4cm,height=1.7cm]{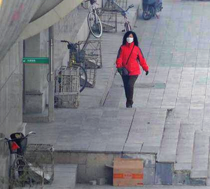}
  \includegraphics[width=2.4cm,height=1.7cm]{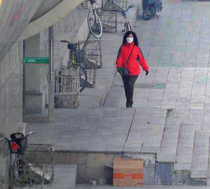}
  \includegraphics[width=2.4cm,height=1.7cm]{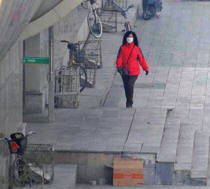}
  \includegraphics[width=2.4cm,height=1.7cm]{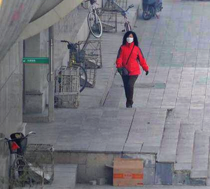}
  \includegraphics[width=2.4cm,height=1.7cm]{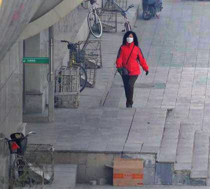}
  \includegraphics[width=2.4cm,height=1.7cm]{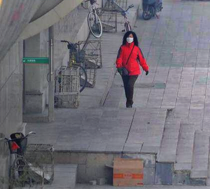}
  \\
  \leftline{\hspace{1.0cm}  (h) 0.49 \hspace{1.0cm}   (i) 0.50 \hspace{1.2cm}  (j) 0.51 \hspace{1.2cm} (k) 0.52 \hspace{1.2cm}  (l) 0.53  \hspace{1.2cm} (m) 0.54 \hspace{1.0cm} (n) 0.56}
  \caption{Visual dehazing results with different controllable parameters on RTTS~\cite{li2018benchmarking}.}
  \label{fig:diff_alpha_RTTS}
\end{figure*}

\IEEEPARstart{H}{aze} is a natural phenomenon that causes the quality of the image to decrease as the haze concentration increases. The purpose of dehazing task is to recover clear scene content from hazy images. In order to accurately remove the haze in the hazy image and restore the scene information as completely as possible, image dehazing has been explored by many recent studies~\cite{wu2021contrastive,kim2021pixel,Li2021deep_retinex,li2021yoly,shyam2021towards,zhang2021hierarchical,zhang2021hierarchical_tcsvt,yin2021visual,li2022physically,liu2022towards}. Existing studies have achieved impressive image restoration performance in terms of providing a single dehazed image during the inference stage. Here, we define the research scope of single-output and multi-output dehazing models from the perspective of users. A single-output dehazing network provides a single dehazed result, while a multi-output dehazing network can output different dehazed images at the inference stage. Thus, the division of single-output and multi-output here is based on whether the inference stage can provide users with a variety of dehazed results, regardless of the behavior of the training process. The research in this paper expands the conventional research goal, that is, to generate multi-output by introducing the illumination controllable ability, and the visual effects are shown in Fig. \ref{fig:diff_alpha_ITS}, Fig. \ref{fig:diff_alpha_OTS} and Fig. \ref{fig:diff_alpha_RTTS}. 

The atmospheric scattering model~\cite{mccartney1976optics,nayar1999vision} is a physical model widely used in image dehazing algorithms. It describes the relationship between the hazy image $I(x)$ and the corresponding clear image $J(x)$ as following
\begin{equation}
  \label{eq:asm}
  I(x) = J(x)t(x) + A(1-t(x)),
\end{equation}
where $A$ and $t(x)$ represent global atmospheric light and  transmission map, respectively. Since the formation of haze can be approximated by atmospheric scattering model,~\cite{he2010single,zhu2015fast} employ physical priors to estimate the parameters required for the dehazing process. However, physical priors are not always valid and they may work well on some images but not others. Meanwhile, when Convolutional Neural Networks (CNNs)~\cite{gui2022comprehensive} are explored in computer vision research, various deep learning-based supervised dehazing methods~\cite{gui2022comprehensive} are proposed. Different network construction strategies~\cite{hong2020dist,zhang2021semantic,qin2020FFA} in CNNs are applied to the image dehazing architectures.

Existing studies~\cite{engin2018cycle,li2020zero,wang2022cycle}, have explored a few effective unsupervised dehazing methods. The perceptual loss is used by CycleDehaze~\cite{engin2018cycle} to optimize the dehazing network. The zero-shot learning and atmospheric scattering model are combined by ZID~\cite{li2020zero} and YOLY~\cite{li2021yoly}. USID-Net~\cite{li2022usid} introduces attention mechanism and disentangled representation into the adversarial training process. Deep DCP~\cite{alona2020un_dcp} designs an energy function suitable for unsupervised dehazing. D4~\cite{yang2022self} designs a method for estimating the scattering coefficient and depth information for the dehazing process. SNSPGAN~\cite{wang2022cycle} proposes a cyclic self-perceptual loss which is computed from unpaired data. However, similar to the supervised methods, these unsupervised dehazing methods also lack researches of the multi-output. As stated by ~\cite{li2021dehazeflow}, the dehazing task is an ill-posedness problem, which means that single-output may not be suitable. 
Here, we think there are two issues that need to be considered as follows.
\begin{itemize}
  \item For a specific scene in the real world, there will be a certain time interval between the formation of hazy and haze-free image. During this time interval, the lighting will change. Therefore, for a given hazy image, the dehazing algorithm should not only output a single dehazed result with fixed illumination condition.
  \item If the dehazing algorithm can output different dehazed images, the generating process should be interpretable.
\end{itemize}

Considering the above two problems, this paper proposes a model by using the Retinex model~\cite{land1977retinex,land1971lightness,xue2016video,fan2017improved} to control illumination. On one hand, the training process of the IC-Dehazing utilizes the data learning ability of deep neural networks and prior statistical knowledge. The Swin-Transformer~\cite{liu2021swin} block and prior knowledge~\cite{he2010single} are embedded into Retinex-based dehazing network to obtain high-quality dehazed outputs. On the other hand, using the Retinex model to obtain different outputs is sufficiently interpretable. By choosing different lighting control parameters in the inference stage, IC-Dehazing obtains a variety of user-acceptable outputs as shown in Fig. \ref{fig:diff_alpha_ITS}, Fig. \ref{fig:diff_alpha_OTS} and Fig. \ref{fig:diff_alpha_RTTS}. The main difference among dehazed images of the same scene is the effect of illumination intensity on scene reflections, such as the table in Fig. \ref{fig:diff_alpha_ITS}, the sky in Fig. \ref{fig:diff_alpha_OTS} and the road in Fig. \ref{fig:diff_alpha_RTTS}.

 
Overall, the main contributions of this paper are listed as follows.
\begin{itemize}
  \item Considering that the illumination information of clear images can not be uniquely determined, an illumination controllable dehazing network is proposed, called IC-Dehazing. The core component of IC-Dehazing is the illumination controllable module. Selecting different illumination weights during the inference phase can provide various outputs, which represent a variety of high-quality dehazed images.
  \item Using dark channel prior~\cite{he2010single} to construct pseudo-labels, a pre-trained model-based loss that decays with the number of iterations is designed, which can promote the performance of the dehazing network.
  \item The IC-Dehazing with the Transformer architecture can be trained in a completely unsupervised manner, which means that pairs of hazy and haze-free images are not required. The research in this paper shows that Transformer can be embedded into the unsupervised dehazing process. Ablation experiments show that Transformer-based unsupervised architecture performs better on dehazing task than conventional convolution-based unsupervised architecture.
\end{itemize}

To verify the performance of IC-Dehazing, extensive quantitative and qualitative experiments are conducted on a variety of dehazing algorithms. Further, the impact of image dehazing on semantic segmentation and object detection is analyzed. Compared to state-of-the-art unsupervised methods, IC-Dehazing achieves better quantitative and qualitative results. As for state-of-the-art supervised algorithms, IC-Dehazing also achieves competitive performance.

\section{Related Work}
Image dehazing algorithms can be roughly divided into two types: pair-based and non-pair-based dehazing, which usually mean supervised and unsupervised. pair-based methods usually use supervised CNNs~\cite{qin2020FFA,chen2019Gated,li2017all}, while non-pair-based methods can be based on statistical assumptions~\cite{he2010single,dhara2020color,liu2021joint,zhu2020novel} or unsupervised CNNs~\cite{li2022usid,li2020zero,wang2022cycle,cong2020discrete}. This section will review researches on dehazing, both pair-based and non-pair-based algorithms. Further, multi-output dehazing and Retinex theory are investigated to illustrate why the IC-Dehazing is proposed.


\subsection{Pair-based Dehazing}
Dehazing methods based on paired data usually employ a supervised learning strategy, which usually use ground-truth clear images or transmission maps as labels. Different powerful network structures~\cite{qin2020FFA,chen2019Gated} are proposed for feature extraction and clean image reconstruction. 4KDehazing~\cite{zheng2021ultra} using bilateral grid learning designs a fast dehazing model for high-resolution images. Guided by contrastive learning, AECR-Net~\cite{wu2021contrastive} proposed a contrastive regularization strategy to improve the similarity between dehazed images and ground-truth labels. MSBDN~\cite{dong2020multi} introduces boosting and error feedback into the training process, which improves the ability to recover spatial information. 

Transformer-based architectures have been applied to image denoising~\cite{zamir2022restormer,wang2022uformer}, object detection~\cite{dong2022cswin}, and image segmentation~\cite{liu2021swin} tasks. Meanwhile, there are relatively few studies on Transformer-based dehazing. DeHamer~\cite{guo2022image} combines CNNs and Transformer, which adds CNNs encoder and decoder to the overall dehazing process. AIDTransformer~\cite{Kulkarni_2023_WACV} utilizes deformable multi-head attention with spatially attentive offset extraction based solution for the designing of the dehazing architecture. The recently proposed DehazeFormer~\cite{song2022vision} uses the Swin-Transformer~\cite{liu2021swin} to build the U-shape architecture. DehazeFormer proved that cyclic shift is not suitable for dehazing problem. It uses reflection padding instead of cyclic shift, which IC-Dehazing follows. The IC-Dehazing proposed in this paper also uses Swin-Transformer to build a dehazing network. However, IC-Dehazing and DehazeFormer have three notable differences as follows. First, unlike DehazeFormer that has a single dehazed result by using the atmospheric scattering model, IC-Dehazing's dehazing module contains two decoded outputs for Retinex fusing, which are designed to control illumination information. Second, DehazeFormer utilizes various improvements, including SK~\cite{li2019selective} fusion, soft reconstruction, rescale layer normalization, additional convolution, improved nonlinear activation function, which are not used in IC-Dehazing.
Third, DehazeFormer proves that Transformer can be used for supervised dehazing training, whereas IC-Dehazing proves that Transformer can be embedded into unsupervised dehazing training process.


\subsection{Non-pair-based Dehazing}
Research that is not based on paired data mainly includes methods based on physical priors and methods based on deep learning, or utilizes both. He~\cite{he2010single} proposed the dark channel prior (DCP) based on the statistics experience of the image, which inspired the research of unsupervised dehazing methods based on deep learning. Golts et al.~\cite{alona2020un_dcp} designed Deep DCP on the basis of DCP, which is an unsupervised model that can be directly optimized in an end-to-end manner. Mo et al.~\cite{mo2021dca} utilizes the dark channel loss as an auxiliary loss term to guide model optimization, which can obtain higher quality generated image. The IC-Dehazing proposed in this paper also uses the DCP to construct pseudo-label for optimizing the dehazing model. Unlike~\cite{alona2020un_dcp} and~\cite{mo2021dca}, IC-Dehazing adopts a feature-based dark channel decay loss based on a pre-trained model to provide the guidance for dehazing network.

PDI2A~\cite{li2022physically} proposes a physically disentangled joint intra- and inter-domain adaptation paradigm, which adopts a semi-supervised training process. USID-Net~\cite{li2022usid} designs multi-scale feature attention and maps the data into the latent space in a shared form. In USID-Net, unsupervised dehazing is achieved with the help of an adversarial training process. D4~\cite{yang2022self} proposes a method for estimating the scattering coefficient and depth information contained in hazy and clean images during the training of the dehazing network. SNSPGAN~\cite{wang2022cycle} develops a cyclic self-perceptual loss that can be computed from unpaired data. ZID~\cite{li2020zero} and YOLY~\cite{li2021yoly} simultaneously utilize zero-shot learning, unsupervised learning and atmospheric scattering models to design an end-to-end dehazing model that do not require training data. Although they reduce the model's reliance on large amounts of data, they require hundreds of iterations to train for each new hazy image.

\subsection{Multi-output Dehazing and Retinex Theory}
\label{subsec:retinex_multioutputs_dehazing}
The quantitative evaluation of image dehazing algorithms is based on the ground-truth image~\cite{ancuti2016d}, so the hazy image and the ground-truth are taken ``simultaneously''. Therefore, for the quantitative evaluation of dehazing task~\cite{qin2020FFA}, the observer can clearly indicate whether the dehazed image is close to the ground truth. In the real-world application, the observer cannot take pictures of hazy and haze-free images ``simultaneously''. Haze affects the reflection process of light, changing the illumination intensity of the image, thus causing color distortion, contrast reduction and brightness imbalance in the image. Given a hazy image, since dehazing is an ill-posedness problem~\cite{li2021dehazeflow}, the color and brightness of the image after dehazing cannot be uniquely identified. Therefore, we argues that in real-world application, the algorithm should fully consider the control of illumination. Specifically, the dehazing results for different illumination values should be presented to the observer.

To the best of our knowledge, two multi-output dehazing models have been proposed, that are pWAE~\cite{kim2021pixel} and DehazeFlow~\cite{li2021dehazeflow}. The core network of pWAE is pixel-wise Wasserstein autoencoders~\cite{kim2021pixel}, which uses the idea of style transfer. The purpose of DehazeFlow is to learn the conditional distribution of a clear image from a hazy image, and its main contribution is to use normalizing flow for image dehazing tasks. For DehazeFlow, the mapping of its training process is the conditional distribution from clean images to noise, while the inference process is the opposite. Unlike pWAE and DehazeFlow, the proposed IC-Dehazing is a combination of physical model and unsupervised neural network, that is, using Retinex theory, dark channel prior and unsupervised Transformer together. Therefore, compared to pWAE and DehazeFlow, which lack physical interpretability, the different outputs of IC-Dehazing clearly represent different lighting conditions.

As mentioned earlier, the illumination controllability of IC-Dehazing is based on the Retinex theory. Galdran~\cite{galdran2018duality} proposed that the atmospheric scattering model and Retinex have a duality under the assumption that the atmospheric light value is a fixed constant. With the help of model duality, Retinex is successfully used for non-data-driven dehazing tasks. Compared with the widely used atmospheric scattering model~\cite{mccartney1976optics}, the application of Retinex theory to deep neural network dehazing~\cite{Li2021deep_retinex} is relatively less explored.
\section{Methods}
\label{sec:methods}
To be able to control the illumination of the dehazed image, a model for Retinex fusion is proposed. The dehazing task requires the generated restored images to be of high quality. Thus, the Transformer-based encoder-decoder network is designed for the dehazing process. Meanwhile, the acquisition of real-world pairwise hazy and haze-free data is relatively difficult, so IC-Dehazing is designed in an unsupervised manner, which is inspired by CycleGAN~\cite{zhu2017unpaired}. Since the training process of unsupervised illumination controllable is difficult, a loss based on dark channel prior is proposed to assist the initial training of the model. This section will introduce the details of the dehazing algorithm, including the illumination controllable Transformer, the loss module with prior guidance, the unsupervised training framework, and the overall loss function. $I$ and $J$ represent the hazy and haze-free domain, where $x \in I$ and $y \in J$.
\subsection{Illumination Controllable Module with Retinex}
\label{subsec:CI_module}

\begin{figure*}
  \centering
  \includegraphics[width=17.39cm,height=5cm]{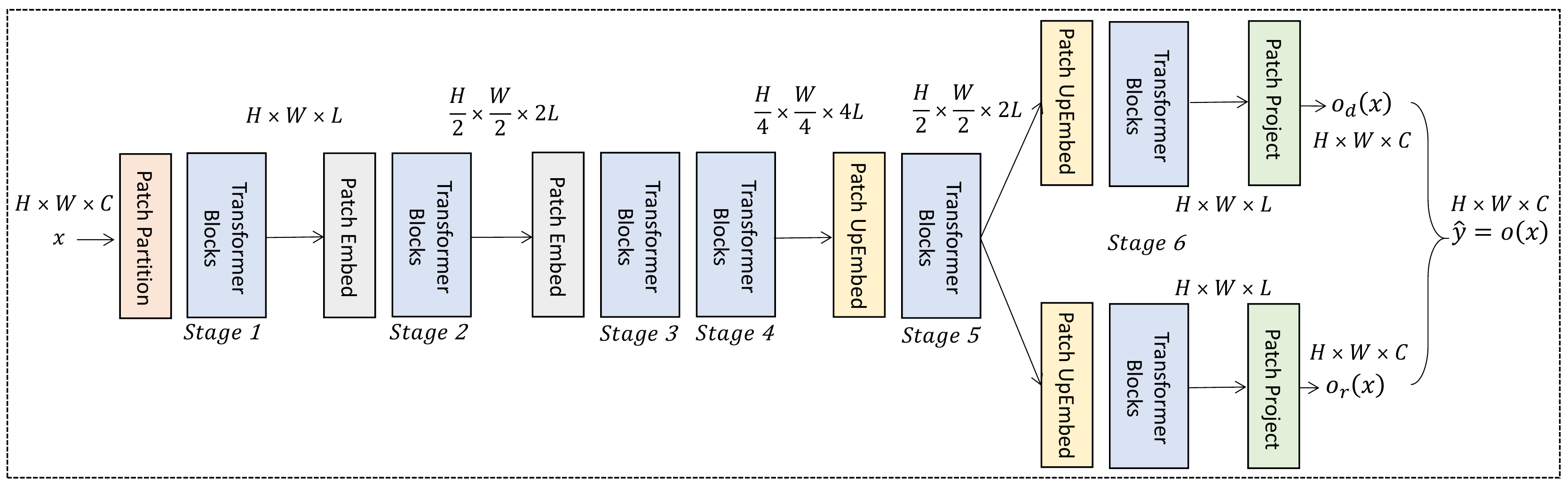}
  \caption{Network details of CIDM.}
  \label{fig:CIDM}
\end{figure*}

\begin{figure*}
  \centering
  \includegraphics[width=17.39cm,height=4cm]{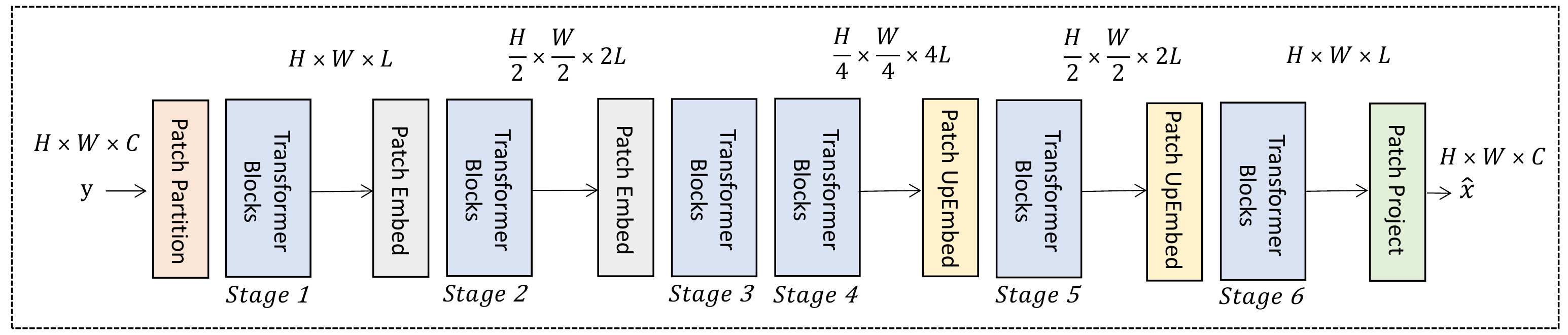}
  \caption{Network details of HSM.}
  \label{fig:HSM}
\end{figure*}

The output of dehazing should not be unique since we do not know the specific lighting conditions of a haze-free scene. Based on the Retinex theory, a continuously controllable module of the illumination map is designed.~\cite{Li2021deep_retinex} uses the Retinex model for image dehazing, which proposes that hazy image $x$ and the dehazed image $\hat{y}$ can be described by the following model
\begin{equation}
  x = R * L_{h},
  \hat{y} = R * L_{dh},
\end{equation}
where $L_{h}$ and $L_{dh}$ are the illumination maps corresponding to $x$ and $\hat{y}$, respectively. The $*$ is element-wise multiplication. The reflectance $R$ is illumination invariant. Therefore, the following relationship can be established between the hazy image and the dehazed image
\begin{equation}
  \label{eq:retinex}
  x = R * L_{dh} * \frac{L_h}{L_{dh}} = \hat{y} * \frac{L_h}{L_{dh}}.
\end{equation}
Therefore, a hazy image can be obtained by the formula, that is, $x = \hat{y} * L_{r}$, where $L_{r} = \frac{L_h}{L_{dh}}$.~\cite{Li2021deep_retinex} calls $L_{r}$ a residual illumination map. Here, we fuse Retinex mapping and direct mapping together, instead of using Eq. (\ref{eq:retinex}) directly.

For a given input $x$, the Transformer encoding module used for feature extraction is denoted as $\phi(\cdot)$, and the corresponding intermediate feature map is $\phi(x)$. Then, two decoders are used to learn direct mapping and Retinex mapping. The first decoding function $\varphi_{d}(\cdot)$ is
\begin{equation}
  o_{d}(x) = \varphi_{d}(\phi(x)).
\end{equation}
The encoding and decoding paths above are used to obtain a direct mapping from a hazy image to a haze-free image. As described above, a hazy image can be described as $x = \hat{y} * L_{r}$ according to Retinex theory. Therefore, a dehazed image can be obtained by $\hat{y} = x * L_{r}^{-1}= x * L_{ir}$, where the inverted residual illumination map $L_{ ir}=L_{r}^{-1}$ (For convenience, $L_{ir}$ is hereinafter also referred to as the illumination map). The second decoding function $\varphi_{r}(\cdot)$ based on the Retinex model is
\begin{equation}
  o_{r}(x) = \varphi_{r}(\phi(x)) * {x} = L_{ir} * x.
\end{equation}
$L_{ir} = \varphi_{r}(\phi(x))$ is learned by Transformer network, based on the Retinex theory that can describe the human visual perception system.

Finally, the direct decoding and the Retinex decoding layer are linearly fused. Meanwhile, the output is mapped to $(-1, 1)$ by the hyperbolic tangent function as
\begin{equation}
  \label{eq:fusion_of_od_and_of}
  o_f(x) = \alpha \times o_{d}(x) + (1 - \alpha) \times o_{r}(x),
\end{equation}
\begin{equation}
  \label{eq:CI_module}
  o(x) = \frac{e^{o_f{(x)}} - e^{-o_f{(x)}}}{e^{o_f{(x)}} + e^{-o_f{(x)}}}.
\end{equation}

Therefore, the final result of dehazing model is a linear superposition of the two parts. The training phase of IC-Dehazing includes the reconstruction process of $x \rightarrow \hat{y} \rightarrow x_{r}$ as shown in Fig. \ref{fig:overall_training_process}. Therefore, $\alpha$ should be set to a fixed value during the training process to ensure the accuracy of reconstruction process. In the inference stage, different $\alpha$ values can be selected from an interval to obtain the output of different illuminations. For convenience, we refer to the fusion process represented by Eq. (\ref{eq:fusion_of_od_and_of}) and Eq. (\ref{eq:CI_module}) as CI Module.

\subsection{Overall Network Structure}
\label{subsec:overall_network}
IC-Dehazing consists of an illumination controllable dehazing module (CIDM), a haze synthesis module (HSM) and two discriminators. The backbone network corresponding to the CIDM and HSM comes from Swin-Transformer~\cite{liu2021swin}. Both CIDM and HSM are hierarchical Transformer architectures with 6 stages. Each stage is composed of multiple Transformer Blocks. The overall structure of the CIDM and HSM are shown in Fig. \ref{fig:CIDM} and Fig. \ref{fig:HSM}. 

\subsubsection{CIDM and HSM}
CIDM consists of two downsampling and two upsampling processes. The input and output dimensionalities of CIDM are $H \times W \times C$, where $H$, $W$, and $C$ represent the height, width, and number of channels of the image, respectively. During downsampling, both the height and width of the feature map are reduced to half of the previous layer, namely $\frac{H}{2} \times \frac{W}{2} \times 2L$ and $\frac{H}{4} \times \frac{W}{4} \times 4L$, where $L$ denotes the feature dimensionality. The size of the feature map increases continuously during the upsampling process until it is restored to $H \times W \times C$. Both CIDM and HSM contain 6 stages and the difference is that CIDM contains CI Module that HSM does not have. Stages 1-4 of CIDM and HSM represent the encoding process, and the features of the image are extracted through the Transformer block. Meanwhile, their Stage 5 and Stage 6 represent the decoding process. In particular, it should be pointed out that the two decoders of CIDM $\varphi_{d}(\cdot)$ and $\varphi_{r}(\cdot)$ share the weight of Stage 5. Through the hierarchical encoding and decoding process, high-quality dehazed image can be obtained by CIDM.

Table \ref{tab:network_config} gives the network parameter configuration of CIDM and HSM. In order to reduce the number of parameters of the network, like~\cite{song2022vision}, several blocks do not add Attention. Specifically, the latter Transformer blocks of each stage use  Attention while the former ones do not. For blocks without Attention, standard convolutions are used for feature extraction. Stage 6 of the CIDM contains two identically structured parts, both consistent with the last row of Table \ref{tab:network_config}.

\begin{table}
  \centering
  \caption{Network configuration of CIDM and HSM.}
  \label{tab:network_config}
  \begin{tabular}{ccccc}
    \hline
    Stage & Blocks & Embed & Attention & Heads \\
    \hline
    1     &   4    &   24  &    1/4    &   2    \\
    2     &   8    &   48  &    1/2    &   4    \\
    3     &   8    &   96  &    1/2    &   4    \\
    4     &   8    &   96  &    1/2    &   4    \\
    5     &   4    &   48  &    1/4    &   2    \\
    6     &   4    &   24  &    1/4    &   1    \\
    \hline
  \end{tabular}
\end{table}

\subsubsection{Patch Partition, Patch Embed, Patch UpEmbed and Patch Project}
Patch Partition is used to project the image into feature space, which is realized by a single convolution layer with stride 1. Patch Embed and Patch UpEmbed are used for dimensionality reduction and upscaling process, respectively. Since dehazing is a pixel-level image restoration task, standard convolution is used in Patch Embed and Patch UpEmbed to preserve the spatial information of feature maps. For Patch Embed, a single-layer convolution with stride of $2$ is employed. Inspired by ~\cite{liang2021swinir,zamir2022restormer}, pixel shuffle~\cite{shi2016real} is used for the upscaling. For an input $x$ with $m$ channels, assume that the desired number of output channels is $n$. Patch UpEmbed maps the number of channels of $x$ to $n \times 2^{2}$, and then obtains the output of $n$ channels by pixel shuffle with an upscale factor of $2$. The Patch Project is used to obtain the final output, reducing the number of channels from $L$ to $3$ by a single-layer convolution with stride of $1$.

\subsubsection{Transformer Block}
For an input feature map, it is projected to $Q$ (query), $K$ (key) and $V$ (value) within $M \times M$ local windows, where $d$ is the query/key dimensionality and $Q , K, V \in R^{M^{2} \times d}$. Then, the self attention is obtained as
\begin{equation}
  Attention(Q, K, V) = softmax(\frac{QK^{T}}{\sqrt{d}} + B) V,
\end{equation}
where $B$ is the relative positional encoding. The Transformer block consists of window-based multi-head self attention (W-MSA) and multi-layer perceptron (MLP). For an input $z^{l-1}$, it first goes through W-MSA and LayerNorm (LN) ~\cite{liu2021swin} layers. The obtained intermediate feature maps are summed with the original input to form residual connections. Then, the final output $z^{l}$ is calculated using LN and MLP with two fully connected layers, where residual connections are also utilized. The overall calculation process is
\begin{equation}
  \hat{z}^{l} = W-MSA(LN(z^{l - 1})) + z^{l-1},
\end{equation}
\begin{equation}
  z^{l} = MLP(LN(\hat{z}^{l})) + \hat{z}^{l}.
\end{equation}

\subsubsection{Discriminator Networks}
For the discriminator networks, the residual discriminator of ~\cite{zhu2017unpaired} is adopted by the proposed IC-Dehazing. For an input image with 3 channels, we first use convolution with stride 1 to obtain a feature map with 64 channels. Then after 4 convolutions with stride 2 (followed by BatchNorm and LeakyReLU), the feature map size is reduced by half and the number of channels is doubled after each convolution. The final output patch with channel number 1 is obtained by a 3x3 convolution with stride 1. 

\subsection{Loss Module with Prior Guidance}
\label{subsec:dcp_loss}
Since the training process of IC-Dehazing is unsupervised, in order to help Transformer-based CIDM learn the mapping from hazy to haze-free images at the early stage of training, the prior loss $L_{\upvartheta}$ is used to guide CIDM. The parameters of CIDM are randomly initialized at the beginning of training, thus using prior law to guide CIDM can play a warm-up role. As the training process progresses, the weight of $L_{\upvartheta}$ is continuously reduced, so that CIDM mainly learns the dehazing mapping from the data. DCP~\cite{he2010single} has been shown to be effective for dehazing task. For a given image $J$, DCP can be expressed as
\begin{equation}
  J^{dark}(u) = \min_{y \in \Omega{(u)}}(\min_{c \in \{r, g, b\}}J^{c}(y)),
\end{equation}
where $J^{c}$ denotes a color channel of $J$, $\Omega(u)$ means a local patch. $\Omega(u)$ is centered at $u$. Here, a perceptual loss based on DCP is proposed. DCP is denoted as $\upvartheta(\cdot)$, and the corresponding pseudo-label is $\upvartheta(x)$, where $x$ is a hazy image. In order to take advantage of prior knowledge and data-driven strategy simultaneously, the loss function $L_{\upvartheta}(\upvartheta(x),o(x))$ is designed in a decaying form as
\begin{equation}
\label{eq:L_dcp}
 L_{\upvartheta}(\upvartheta(x), o(x)) = \upgamma ||\psi{(\upvartheta(x))} - \psi{(o(x))}||_{2},
\end{equation}
where $\psi(\cdot)$ represents the pretrained classification network with the fully connected layers removed, and VGG19~\cite{K2015vgg} is chosen in this paper. Here we do not directly calculate the distance between $o(x)$ and $\upvartheta(x)$, but calculate the distance between the features of the two. The reason for this is that the purpose of the prior loss is to keep the dehazing result consistent with the regularity of clear image at the feature level. 

The value of $L_{\upvartheta}$ is constantly decaying with iterations, and $\upgamma$ is the decay coefficient for each epoch. It will be set to different values for different datasets and details can be found in the experiments section.

\subsection{Overall Training Process and Loss Functions}
\begin{figure}
  \centering
  \includegraphics[width=5.5cm,height=2.0cm]{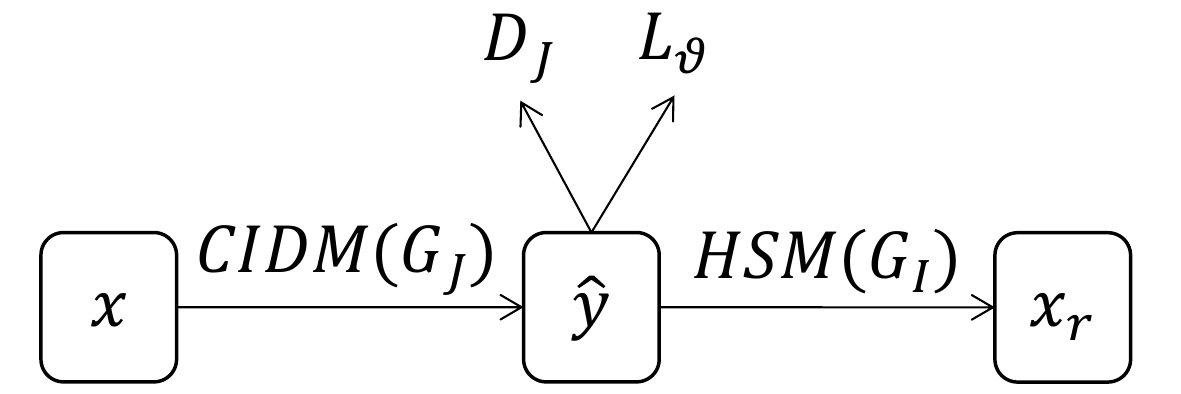}
  \\
  \begin{small}
    (a) The reconstruction process of $x$, which is denoted as P1. 
  \end{small}
  \includegraphics[width=5.5cm,height=2.0cm]{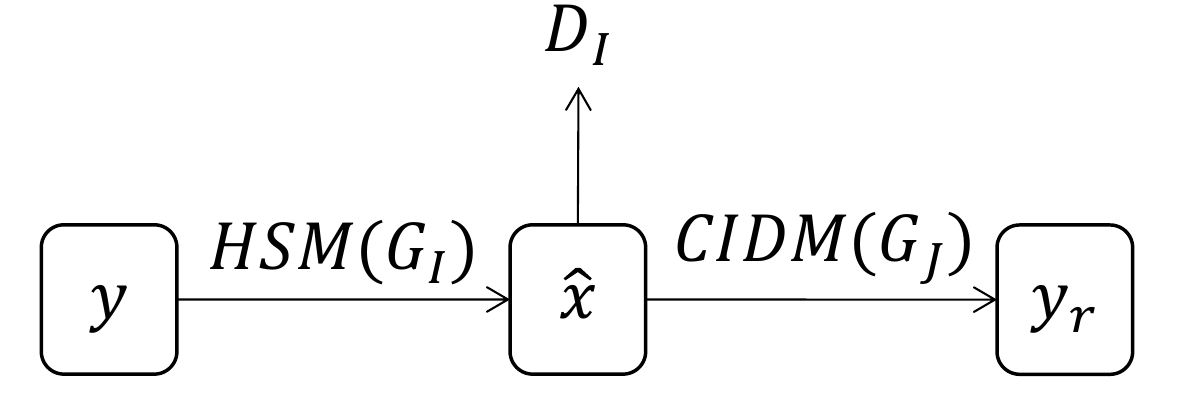}
  \\
  \begin{small}
    (b) The reconstruction process of $y$, which is denoted as P2. 
  \end{small}
  \caption{The overall training process of IC-Dehazing, where $L_{\upvartheta}$ denotes the prior constraint. $x_{r}$ and $y_{r}$ represent the reconstructed $x$ and $y$, respectively.}
  \label{fig:overall_training_process}
\end{figure}

The training of IC-Dehazing is an unsupervised process that does not require pairwise data. The training method is inspired by~\cite{zhu2017unpaired}. However, unlike~\cite{zhu2017unpaired}, considering the specificity of the dehazing task, IC-Dehazing uses a well-designed prior loss to constrain CIDM, which is introduced in Section \ref{subsec:dcp_loss}. CIDM and HSM are denoted as $G_{J}$ and $G_{I}$, respectively. The two discriminators are marked as $D_{J}$ and $D_{I}$, respectively. As shown in Fig. \ref{fig:overall_training_process}, take P1 as an example, send $x$ to CIDM to obtain dehazed images $\hat{y}$ ($o(x)$ in Fig. \ref{fig:CIDM}), then calculate $L_{\upvartheta}$ and adversarial loss by pre-training model and $D_{J}$, respectively. Further, put $\hat{y}$ to the HSM to obtain the reconstructed images $x_{r}$. For P2, the difference from P1 is that there is no $L_{\upvartheta}$. Therefore, parameter for CIDM and HSM are updated by adversarial loss, prior loss and cycle consistency loss. The adversarial loss is

\begin{small}
\begin{equation}
\begin{aligned}
  L(G_{J}, D_J) = E_{y \sim p_y}[(D_{J}(y) - 1)^2] + E_{x \sim p_x}[D_{J}(G_{J}(x))^2],
\end{aligned}
\end{equation}
\end{small}
\begin{small}
\begin{equation}
  \begin{aligned}
    L(G_{I}, D_I) = E_{x \sim p_{x}}[(D_{I}(x) - 1)^2] + E_{y \sim p_{y}}[D_{I}(G_{I}(y))^2].
  \end{aligned}
  \end{equation}
\end{small}
In order to ensure that the content of the scene during reconstruction remains unchanged, the cycle consistency loss and identity loss~\cite{zhu2017unpaired} are used
\begin{equation}
  \begin{aligned}
    L_{r}(G_{I}, G_{J}) &= E_{x \sim p_{x}}[||G_{I}(G_{J}(x)) - x||_2] + \\
                  &E_{y \sim p_{y}}[||G_{J}(G_{I}(y)) - y||_2] + \\
                  &E_{x \sim p_{x}}[||G_{I}(x) - x||_2] + \\
                  &E_{y \sim p_{y}}[||G_{J}(y) - y||_2].
  \end{aligned}
\end{equation}

A prior-based decay loss function is designed in Section \ref{subsec:dcp_loss}. Therefore, the overall loss function is the sum of adversarial loss, cycle consistent loss and prior loss:
\begin{equation}
  L_{all} = L({G_{J}, D_{J}}) + L({G_{I}, D_{I}}) + L_{r}(G_{I}, G_{J}) + \lambda L_{\upvartheta},
\end{equation}
where $\lambda$ is the initial weight of $L_{\upvartheta}$. It is worth noting that the training of IC-Dehazing is not completely ``symmetric''. For the dehazing process, its output is not only assisted by adversarial training, but also constrained by the prior, since the generated images should be in accordance with the statistical laws of natural haze-free images. For the process of synthesizing haze, its output is mainly guided by the statistical law of the data learned by the adversarial process.

\section{Experiments}
To demonstrate the effectiveness of the proposed IC-Dehazing, quantitative and qualitative experimental analyses of multiple baseline algorithms are performed on different datasets. 

\subsection{Baselines and Settings}
In order to compare the performance of different methods intuitively and fairly, we compare our method with Pair-based and Non-pair-based algorithms. Pair-based methods include DeHamer~\cite{guo2022image}, FFA~\cite{qin2020FFA}, GCANet~\cite{chen2019Gated}, GridDehazeNet~\cite{liu2019grid}, MSBDN~\cite{dong2020multi} and 4KDehazing~\cite{zheng2021ultra}. Non-pair-based algorithms include CANCB~\cite{dhara2020color}, CEEF~\cite{liu2021joint}, FME~\cite{zhu2020novel}, CycleDehaze~\cite{engin2018cycle}, ZID~\cite{li2020zero} and SNSPGAN~\cite{wang2022cycle}. The 12th Gen Intel Core i5-12400f and Matlab 2019b are used as the execution environment for non-deep learning algorithms, which are CANCB, CEEF and FME. NVIDIA Tesla V100 32GB and PyTorch 1.8 framework are used as experimental platforms for deep learning algorithms. 

\subsection{Image Dehazing Evaluation}
This section evaluates the performance of IC-Dehazing from multiple perspectives. The datasets for evaluation are ITS/OTS/RTTS~\cite{li2018benchmarking}, NYU-Seg-Haze~\cite{ancuti2016d} and 4K~\cite{zheng2021ultra}, respectively. A brief introduction is described as follows.
\begin{itemize}
  \item The indoor dataset ITS from RESIDE~\cite{li2018benchmarking}: The training set contains 13,990 pair images and the testing set contains 500 pair images (SOTS indoor). Each hazy image has an accurate ground-truth clear label.
  \item The outdoor dataset OTS from RESIDE~\cite{li2018benchmarking}: The training set contains 72,135 pair images, and the testing set comes from the latest 500 outdoor pair images (SOTS outdoor) provided by the paper website.
  \item 4K~\cite{zheng2021ultra} dataset, which contains high-resolution (3,840 $\times$ 2,160) hazy and haze-free outdoor images. According to the latest published data on the website, 14,605 pair images are selected as the training set and 1,001 pair images are selected as the test set.
  \item Real-world hazy image dataset RTTS without labels, 4,322 in total, is also from ~\cite{li2018benchmarking}.
  \item To evaluate dehazing and semantic segmentation simultaneously, ~\cite{ancuti2016d} is used as one of the dehazing datasets. For convenience, the hazy dataset ~\cite{ancuti2016d} and the corresponding segmentation dataset~\cite{silberman2012indoor} are denoted as NYU-Seg-Haze. A randomly selected 1,399 pair images are used as the training set, and the remaining 100 pair images are used as the test set.
\end{itemize}

The parameter optimization process of all sub-modules of IC-Dehazing adopts the Adam optimizer, where the values of $\beta{1}$ and $\beta{2}$ are 0.9 and 0.999, respectively. The learning rate is 0.0001 for all datasets. Total epochs are 15 (OTS), 25 (ITS/NYU-Seg-Haze) and 45 (4K), respectively. During training, images are randomly cropped to $256 \times 256$ before being put into the network, forming a batch size of 2. The $\upgamma$ is introduced to control the attenuation of the prior loss weight. The choice of $\upgamma$ is different for different datasets, and it is a hyperparameter that needs to be tuned. $\upgamma$ is empirically chosen to be 0.98, 0.95, 0.90, 0.98 for ITS, OTS, 4K, and NYU-Seg-Haze, respectively. In addition, the initial weight value $\lambda$ of $L_{p}(G)$ is fixed to 1. The evaluation metrics for dehazing results are PSNR and SSIM, which are widely used in dehazing tasks~\cite{qin2020FFA,liu2019grid,engin2018cycle,dong2020multi}. Higher PSNR and SSIM values represent better image restoration performance. For RTTS dataset without ground truth clear images, the dehazing results are evaluated using blind image quality assessment metrics NIQE~\cite{mittal2012making} and BRISQUE~\cite{mittal2012no}. Smaller NIQE and BRISQUE values represent better image quality~\cite{wang2022cycle}.

\begin{figure*}
  \centering
  \includegraphics[width=3.4cm,height=2.4cm]{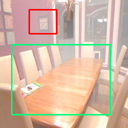}
  \includegraphics[width=3.4cm,height=2.4cm]{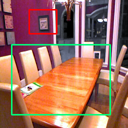}
  \includegraphics[width=3.4cm,height=2.4cm]{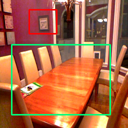}
  \includegraphics[width=3.4cm,height=2.4cm]{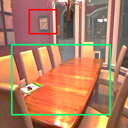}
  \includegraphics[width=3.4cm,height=2.4cm]{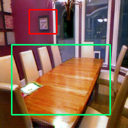}
  \\
  \leftline{\hspace{1.5cm} (a) Hazy \hspace{1.5cm}  (b) CANCB  \hspace{1.7cm}   (c) CEEF  \hspace{2.0cm}  (d) FME  \hspace{1.5cm} (e) CycleDehaze \hspace{1.2cm}  \hspace{1cm}}
  \vspace{0.1cm}
  \includegraphics[width=3.4cm,height=2.4cm]{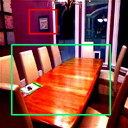}
  \includegraphics[width=3.4cm,height=2.4cm]{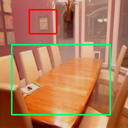}
  \includegraphics[width=3.4cm,height=2.4cm]{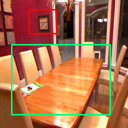}
  \includegraphics[width=3.4cm,height=2.4cm]{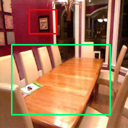}
  \includegraphics[width=3.4cm,height=2.4cm]{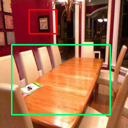}
  \vspace{0.1cm}
  \leftline{\hspace{1.5cm} (f) ZID \hspace{1.7cm}  (g) SNSPGAN \hspace{1.0cm}  (h) IC-Dehazing \hspace{1.1cm}  (i) DeHamer \hspace{1.8cm} (j) FFA \hspace{1.2cm}}
  \\
  \includegraphics[width=3.4cm,height=2.4cm]{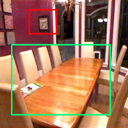}
  \includegraphics[width=3.4cm,height=2.4cm]{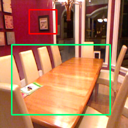}
  \includegraphics[width=3.4cm,height=2.4cm]{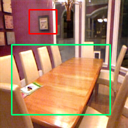}
  \includegraphics[width=3.4cm,height=2.4cm]{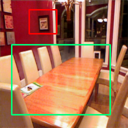}
  \includegraphics[width=3.4cm,height=2.4cm]{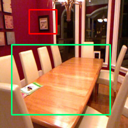}
  \\
  \leftline{\hspace{1.2cm} (k) GCANet \hspace{0.8cm}  (l) GridDehazeNet \hspace{1.0cm}   (m) MSBDN \hspace{1.2cm}  (n) 4KDehazing \hspace{1.5cm} (o) Clear \hspace{1.0cm}}
  \caption{Visual dehazing results on ITS.}
  \label{fig:visual_compare_ITS}
\end{figure*}

\begin{figure*}
  \centering
  \includegraphics[width=3.4cm,height=2.2cm]{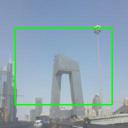}
  \includegraphics[width=3.4cm,height=2.2cm]{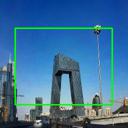}
  \includegraphics[width=3.4cm,height=2.2cm]{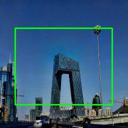}
  \includegraphics[width=3.4cm,height=2.2cm]{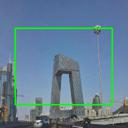}
  \includegraphics[width=3.4cm,height=2.2cm]{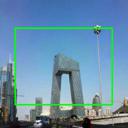}
  \\
  \leftline{\hspace{1.5cm} (a) Hazy \hspace{1.5cm}  (b) CANCB  \hspace{1.7cm}   (c) CEEF  \hspace{2.0cm}  (d) FME  \hspace{1.5cm} (e) CycleDehaze \hspace{1.2cm}  \hspace{1cm}}
  \vspace{0.1cm}
  \includegraphics[width=3.4cm,height=2.2cm]{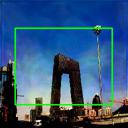}
  \includegraphics[width=3.4cm,height=2.2cm]{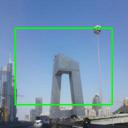}
  \includegraphics[width=3.4cm,height=2.2cm]{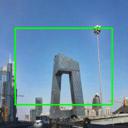}
  \includegraphics[width=3.4cm,height=2.2cm]{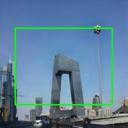}
  \includegraphics[width=3.4cm,height=2.2cm]{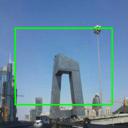}
  \vspace{0.1cm}
  \leftline{\hspace{1.5cm} (f) ZID \hspace{1.7cm}  (g) SNSPGAN \hspace{1.0cm}  (h) IC-Dehazing \hspace{1.1cm}  (i) DeHamer \hspace{1.8cm} (j) FFA \hspace{1.2cm}}
  \\
  \includegraphics[width=3.4cm,height=2.2cm]{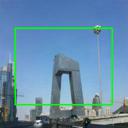}
  \includegraphics[width=3.4cm,height=2.2cm]{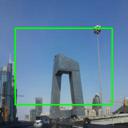}
  \includegraphics[width=3.4cm,height=2.2cm]{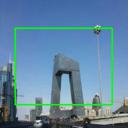}
  \includegraphics[width=3.4cm,height=2.2cm]{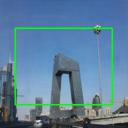}
  \includegraphics[width=3.4cm,height=2.2cm]{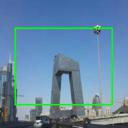}
  \\
  \leftline{\hspace{1.2cm} (k) GCANet \hspace{0.8cm}  (l) GridDehazeNet \hspace{1.0cm}   (m) MSBDN \hspace{1.2cm}  (n) 4KDehazing \hspace{1.5cm} (o) Clear \hspace{1.0cm}}
  \caption{Visual dehazing results on OTS.}
  \label{fig:visual_compare_OTS}
\end{figure*}

\begin{figure*}
  \centering
  \includegraphics[width=3.4cm,height=2.2cm]{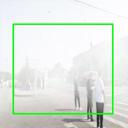}
  \includegraphics[width=3.4cm,height=2.2cm]{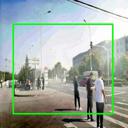}
  \includegraphics[width=3.4cm,height=2.2cm]{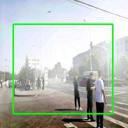}
  \includegraphics[width=3.4cm,height=2.2cm]{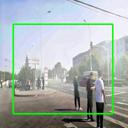}
  \includegraphics[width=3.4cm,height=2.2cm]{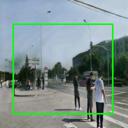}
  \\
  \leftline{\hspace{1.5cm} (a) Hazy \hspace{1.5cm}  (b) CANCB  \hspace{1.7cm}   (c) CEEF  \hspace{2.0cm}  (d) FME  \hspace{1.5cm} (e) CycleDehaze \hspace{1.2cm}  \hspace{1cm}}
  \vspace{0.1cm}
  \includegraphics[width=3.4cm,height=2.2cm]{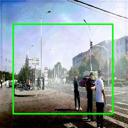}
  \includegraphics[width=3.4cm,height=2.2cm]{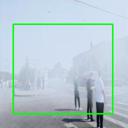}
  \includegraphics[width=3.4cm,height=2.2cm]{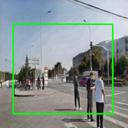}
  \includegraphics[width=3.4cm,height=2.2cm]{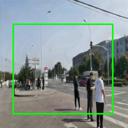}
  \includegraphics[width=3.4cm,height=2.2cm]{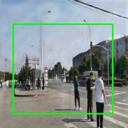}
  \vspace{0.1cm}
  \leftline{\hspace{1.5cm} (f) ZID \hspace{1.7cm}  (g) SNSPGAN \hspace{1.0cm}  (h) IC-Dehazing \hspace{1.1cm}  (i) DeHamer \hspace{1.8cm} (j) FFA \hspace{1.2cm}}
  \\
  \includegraphics[width=3.4cm,height=2.2cm]{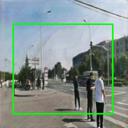}
  \includegraphics[width=3.4cm,height=2.2cm]{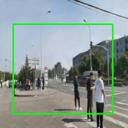}
  \includegraphics[width=3.4cm,height=2.2cm]{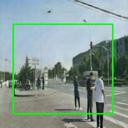}
  \includegraphics[width=3.4cm,height=2.2cm]{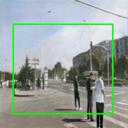}
  \includegraphics[width=3.4cm,height=2.2cm]{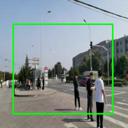}
  \\
  \leftline{\hspace{1.2cm} (k) GCANet \hspace{0.8cm}  (l) GridDehazeNet \hspace{1.0cm}   (m) MSBDN \hspace{1.2cm}  (n) 4KDehazing \hspace{1.5cm} (o) Clear \hspace{1.0cm}}
\caption{Visual dehazing results on 4K.}
\label{fig:visual_compare_4K}
\end{figure*}

\begin{figure*}
  \centering
  \includegraphics[width=3.4cm,height=2.2cm]{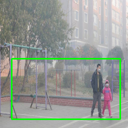}
  \includegraphics[width=3.4cm,height=2.2cm]{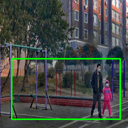}
  \includegraphics[width=3.4cm,height=2.2cm]{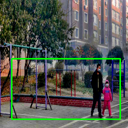}
  \includegraphics[width=3.4cm,height=2.2cm]{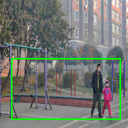}
  \includegraphics[width=3.4cm,height=2.2cm]{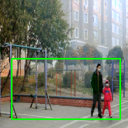}
  \\
  \vspace{0.1cm}
  \includegraphics[width=3.4cm,height=2.2cm]{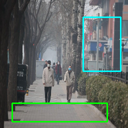}
  \includegraphics[width=3.4cm,height=2.2cm]{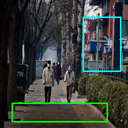}
  \includegraphics[width=3.4cm,height=2.2cm]{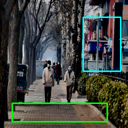}
  \includegraphics[width=3.4cm,height=2.2cm]{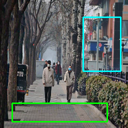}
  \includegraphics[width=3.4cm,height=2.2cm]{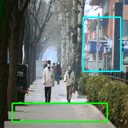}
  \\
   \leftline{\hspace{1.5cm} (a) Hazy \hspace{1.5cm}  (b) CANCB  \hspace{1.7cm}   (c) CEEF  \hspace{2.0cm}  (d) FME  \hspace{1.5cm} (e) CycleDehaze \hspace{1.2cm}  \hspace{1cm}}
  \vspace{0.1cm}
  \includegraphics[width=3.4cm,height=2.2cm]{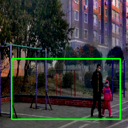}
  \includegraphics[width=3.4cm,height=2.2cm]{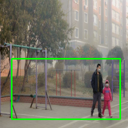}
  \includegraphics[width=3.4cm,height=2.2cm]{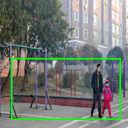}
  \includegraphics[width=3.4cm,height=2.2cm]{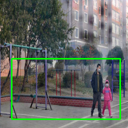}
  \includegraphics[width=3.4cm,height=2.2cm]{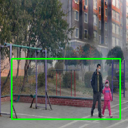}
  \\
  \vspace{0.1cm}
  \includegraphics[width=3.4cm,height=2.2cm]{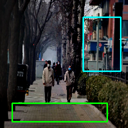}
  \includegraphics[width=3.4cm,height=2.2cm]{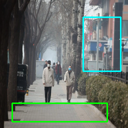}
  \includegraphics[width=3.4cm,height=2.2cm]{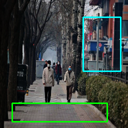}
  \includegraphics[width=3.4cm,height=2.2cm]{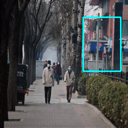}
  \includegraphics[width=3.4cm,height=2.2cm]{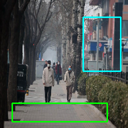}
  \\
  \leftline{\hspace{1.5cm} (f) ZID \hspace{1.7cm}  (g) SNSPGAN \hspace{1.0cm}  (h) IC-Dehazing \hspace{1.1cm}  (i) DeHamer \hspace{1.8cm} (j) FFA \hspace{1.2cm}}
  \vspace{0.1cm}
  \includegraphics[width=3.4cm,height=2.2cm]{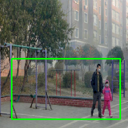}
  \includegraphics[width=3.4cm,height=2.2cm]{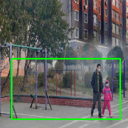}
  \includegraphics[width=3.4cm,height=2.2cm]{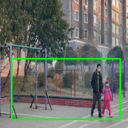}
  \includegraphics[width=3.4cm,height=2.2cm]{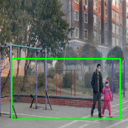}
  \\
  \vspace{0.1cm}
  \includegraphics[width=3.4cm,height=2.2cm]{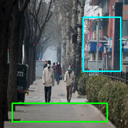}
  \includegraphics[width=3.4cm,height=2.2cm]{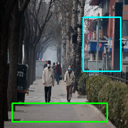}
  \includegraphics[width=3.4cm,height=2.2cm]{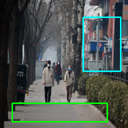}
  \includegraphics[width=3.4cm,height=2.2cm]{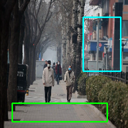}
  \\
  \leftline{\hspace{2.8cm} (k) GCANet \hspace{1.0cm}  (l) GridDehazeNet \hspace{1.0cm}   (m) MSBDN \hspace{1.0cm}  (n) 4KDehazing \hspace{1.0cm}}
\caption{Visual dehazing results on RTTS.}
\label{fig:visual_compare_RTTS}
\end{figure*}

The PSNR and SSIM values obtained by all dehazing algorithms are given in Table \ref{tab:results_on_dehazing}. The results of ZID/Deep DCP~\cite{alona2020un_dcp} on ITS and OTS are consistent with~\cite{li2022usid}. The results of FFA/GridDehazeNet/GCANet/MSBDN on ITS and OTS are consistent with ~\cite{song2022vision}. RTTS does not contain pairs of hazy and haze-free images, so it can not be used in a supervised training process. Therefore, for all deep learning-based dehazing methods, the pre-trained models trained on OTS are directly used for RTTS inference. The experimental results in Table \ref{tab:results_on_dehazing} show that the proposed IC-Dehazing can achieve competitive performance compared to supervised dehazing algorithms. In particular, the PSNR and SSIM obtained by IC-Dehazing on the 4K dataset are 23.355 and 0.936, respectively, which are higher than most supervised algorithms. Since IC-Dehazing does not use pairwise supervision information, it is reasonable that its PSNR and SSIM are slightly lower than supervised algorithms on other datasets. Meanwhile, among different unsupervised algorithms, IC-Dehazing achieves the best overall performance. On the ITS, NYU-Seg-Haze, and 4K datasets, IC-Dehazing achieves the highest values for both quality evaluation metrics. Figs. \ref{fig:visual_compare_ITS}-\ref{fig:visual_compare_RTTS} show the visual dehazing effects of ITS, OTS, 4K and RTTS, respectively. Different colored rectangles are used to emphasize certain areas in the dehazed images. The visual results show that IC-Dehazing can restore the details and texture information of the images very well. It can be concluded that IC-Dehazing is better than the Non-pair-based methods, and can obtain the recovery performance close to the Pair-based algorithms.
\begin{table*}
  \centering
  \caption{Quantitative dehazing results obtained by various dehazing algorithms. $\uparrow$ means that the value is directly proportional to the picture quality, while $\downarrow$ is the opposite.}
  \label{tab:results_on_dehazing}
  \begin{tabular}{cc|cc|cc|cc|cc|cc}
    \hline
    \multirow{2}{*}{Category}      & \multirow{2}{*}{Methods} & \multicolumn{2}{c|}{ITS} & \multicolumn{2}{c|}{OTS} & \multicolumn{2}{c|}{NYU-Seg-Haze} & \multicolumn{2}{c|}{4K} & \multicolumn{2}{c}{RTTS}\\
    \cline{3-12}
    & & SSIM$\uparrow$ & PSNR$\uparrow$ & SSIM$\uparrow$ & PSNR$\uparrow$ & SSIM$\uparrow$ & PSNR$\uparrow$ & SSIM$\uparrow$ & PSNR$\uparrow$ & NIQE$\downarrow$ & BRISQUE$\downarrow$ \\
    \hline
    \multirow{8}{*}{Non-pair-based} & CANCB & 0.869 & 19.830 & 0.862 & 17.229 & 0.773 & 14.855 & 0.790 & 15.746 & 4.287 & 33.010 \\
     & CEEF          & 0.770 & 18.187 & 0.789 & 16.589 & 0.692 & 13.409 & 0.706 & 13.612 & \textbf{4.062} & 29.375 \\
     & FME          & 0.790 & 17.461 & 0.869 & 19.667 & 0.738 & 13.755 & 0.795 & 15.694 & 4.124 & 29.529 \\
     & CycleDehaze  & 0.810 & 18.870  & 0.881  & 23.347 & 0.713 & 13.585 & 0.886 & 21.067 & 4.923 & \textbf{16.103} \\
     & ZID          & 0.835 & 19.830  & 0.633  & 13.520 & 0.536 & 12.313 & 0.506 & 12.499 & 5.166 & 35.649 \\
     & Deep DCP      & 0.832 & 19.250  & \textbf{0.933} & 24.080 & -     & -     &   -   &  -  & - & - \\
     & SNSPGAN      & 0.788 & 17.747  & 0.925  & 24.280 & 0.692 & 12.856 & 0.714  & 14.045 & 5.596 & 34.348 \\
     & IC-Dehazing  & \textbf{0.894} & \textbf{22.558}  & 0.929 & \textbf{24.562} & \textbf{0.774} & \textbf{15.867} & \textbf{0.936} & \textbf{23.355} & 4.489 & 25.152 \\
    \hline                                     
    \multirow{6}{*}{Pair-based} & DeHamer & 0.988 & 36.680 & \textbf{0.986} & \textbf{35.180} & \textbf{0.895} & \textbf{23.867} & \textbf{0.957} & \textbf{26.437} & \textbf{5.083} & 35.383 \\
     & FFA           & \textbf{0.989} & \textbf{36.890} & 0.984 & 33.570 & 0.792 & 18.171 & 0.948 & 25.935  & 5.504 & 34.737 \\
     & GCANet        & 0.980 & 30.230 & 0.945 & 29.912 & 0.742 & 17.539 & 0.895 & 23.543 & 5.297 & \textbf{23.719} \\
     & GridDehazeNet & 0.984 & 32.160 & 0.982 & 30.860 & 0.782 & 18.051 & 0.939 & 22.710 & 5.510 & 29.957 \\
     & MSBDN         & 0.985 & 33.670 & 0.982 & 33.480 & 0.754 & 19.346 & 0.914 & 24.081 & 5.407 & 27.363 \\
     & 4KDehazing    & 0.931 & 26.753 & 0.958 & 29.182 & 0.786 & 18.318 & 0.903 & 21.824 & 5.344 & 31.986 \\
  \hline
  \end{tabular}
\end{table*}

\subsection{Real-world Object Detection Evaluation}
On one hand, the existence of haze will cause the objects in the image to become blurred, which affects the visual effect of the image. On the other hand, for object detection tasks, blurred objects may lead to a decrease in detection accuracy. Therefore, the experiments in this section compare the effects of various dehazing algorithms on the object detection task. The detection evaluation data used is real-world task-driven testing set (RTTS) from RESIDE~\cite{li2018benchmarking}, which contains detection box annotation files corresponding to 4,322 real-world hazy images. RTTS contains five classes, that is person, bicycle, car, motorbike and bus. The currently popular and reliable yolov5~\cite{glenn_jocher_2022_6222936} algorithm is selected as the detection model, which is pre-trained on the COCO~\cite{lin2014microsoft} dataset. The metric of detection task is mAP.5, which is computed by the default setting of yolov5. The results of the five class targets on the original hazy image are 0.813, 0.607, 0.734, 0.637 and 0.578, respectively. The corresponding mAP.5 is 0.674.

\begin{figure*}
  \centering
  \includegraphics[width=3.4cm,height=2.2cm]{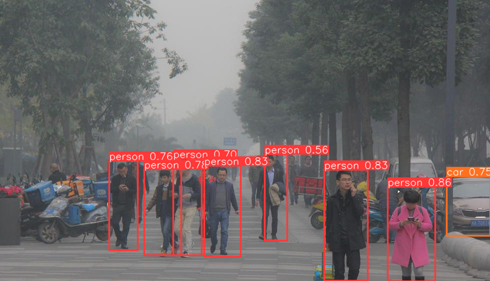}
  \includegraphics[width=3.4cm,height=2.2cm]{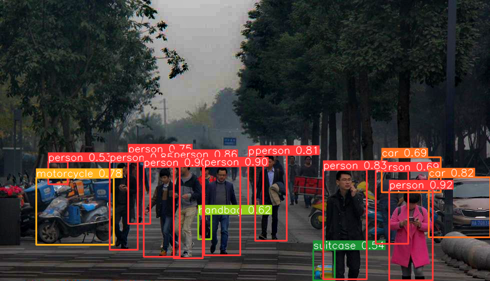}
  \includegraphics[width=3.4cm,height=2.2cm]{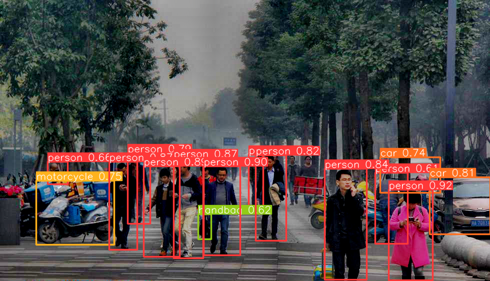}
  \includegraphics[width=3.4cm,height=2.2cm]{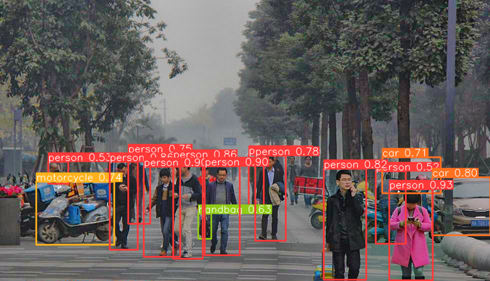}
  \includegraphics[width=3.4cm,height=2.2cm]{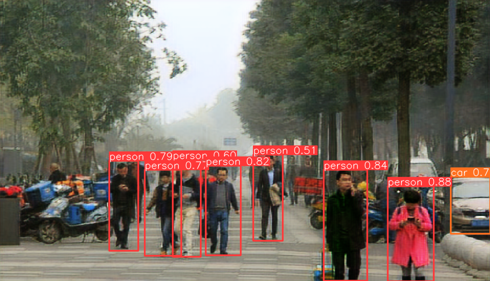}
  \\
  \vspace{0.1cm}
  \includegraphics[width=3.4cm,height=2.2cm]{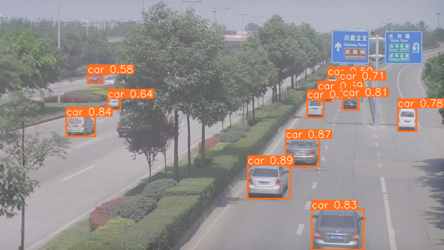}
  \includegraphics[width=3.4cm,height=2.2cm]{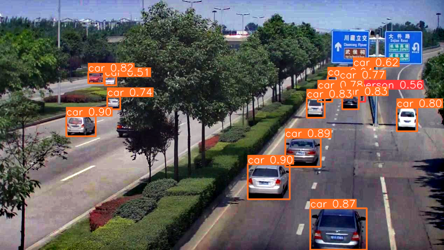}
  \includegraphics[width=3.4cm,height=2.2cm]{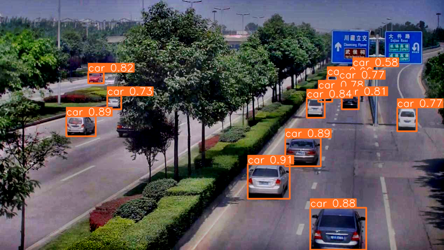}
  \includegraphics[width=3.4cm,height=2.2cm]{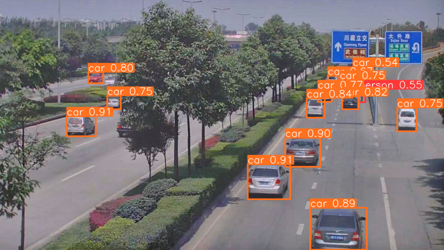}
  \includegraphics[width=3.4cm,height=2.2cm]{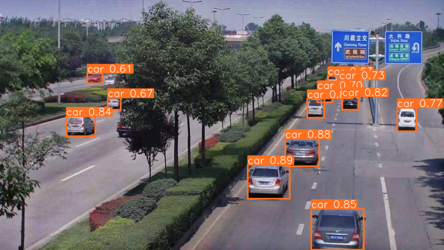}
  \\
  \leftline{\hspace{1.5cm} (a) Hazy \hspace{1.5cm}  (b) CANCB  \hspace{1.7cm}   (c) CEEF  \hspace{2.0cm}  (d) FME  \hspace{1.5cm} (e) CycleDehaze \hspace{1.2cm}  \hspace{1cm}}
  \vspace{0.1cm}
  \includegraphics[width=3.4cm,height=2.2cm]{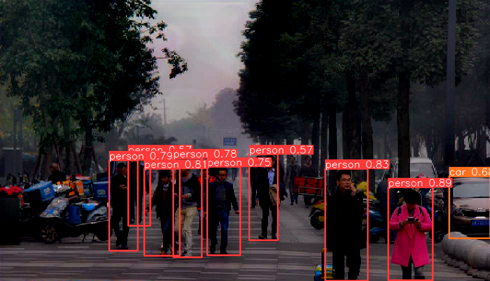}
  \includegraphics[width=3.4cm,height=2.2cm]{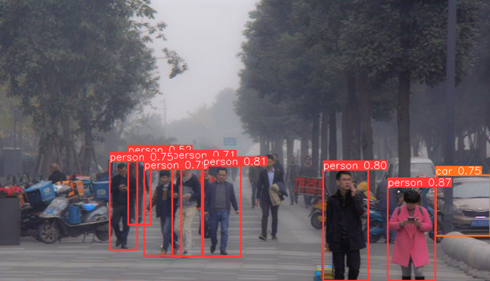}
  \includegraphics[width=3.4cm,height=2.2cm]{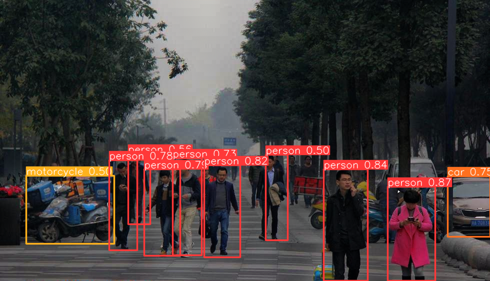}
  \includegraphics[width=3.4cm,height=2.2cm]{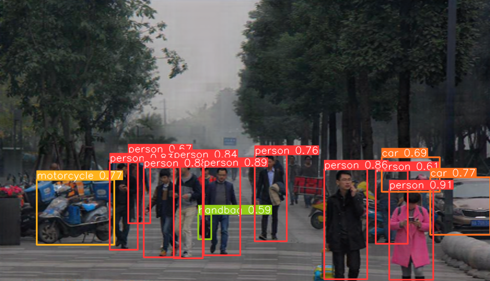}
  \includegraphics[width=3.4cm,height=2.2cm]{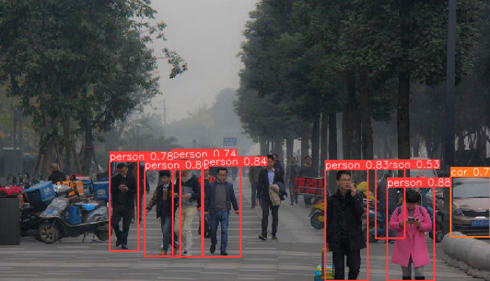}
  \\
  \vspace{0.1cm}
  \includegraphics[width=3.4cm,height=2.2cm]{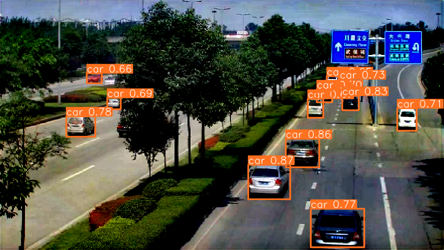}
  \includegraphics[width=3.4cm,height=2.2cm]{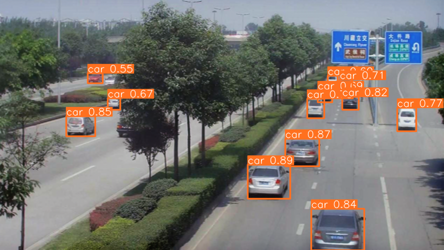}
  \includegraphics[width=3.4cm,height=2.2cm]{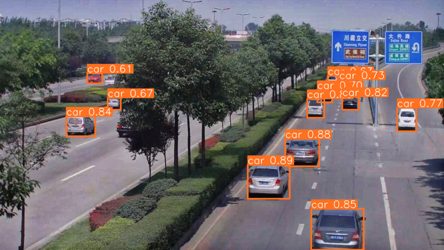}
  \includegraphics[width=3.4cm,height=2.2cm]{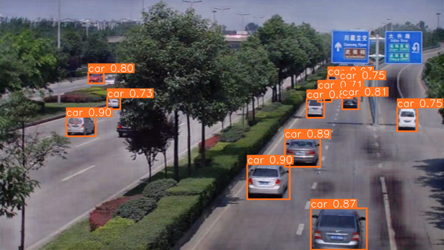}
  \includegraphics[width=3.4cm,height=2.2cm]{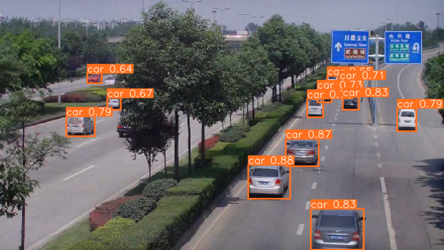}
  \leftline{\hspace{1.5cm} (f) ZID \hspace{1.7cm}  (g) SNSPGAN \hspace{1.0cm}  (h) IC-Dehazing \hspace{1.1cm}  (i) DeHamer \hspace{1.8cm} (j) FFA \hspace{1.2cm}}
  \\
  \vspace{0.1cm}
  \includegraphics[width=3.4cm,height=2.2cm]{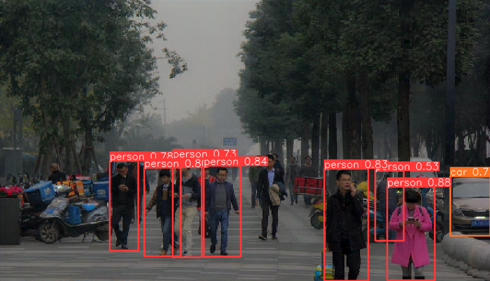}
  \includegraphics[width=3.4cm,height=2.2cm]{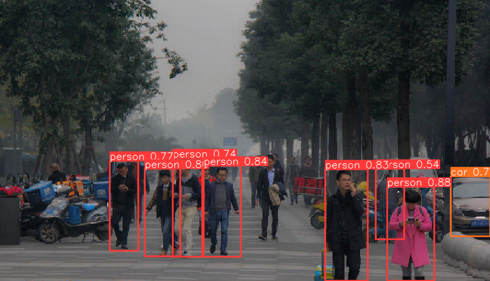}
  \includegraphics[width=3.4cm,height=2.2cm]{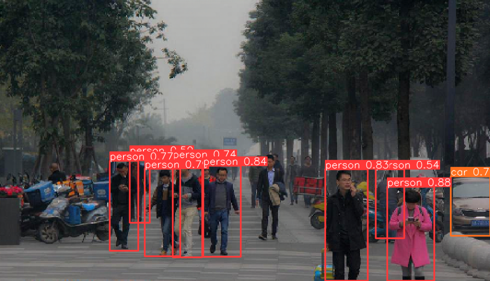}
  \includegraphics[width=3.4cm,height=2.2cm]{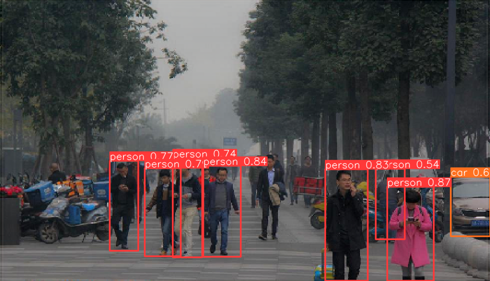}
  \\
  \vspace{0.1cm}
  \includegraphics[width=3.4cm,height=2.2cm]{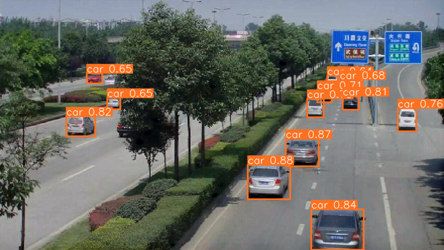}
  \includegraphics[width=3.4cm,height=2.2cm]{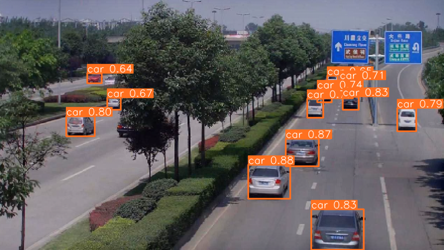}
  \includegraphics[width=3.4cm,height=2.2cm]{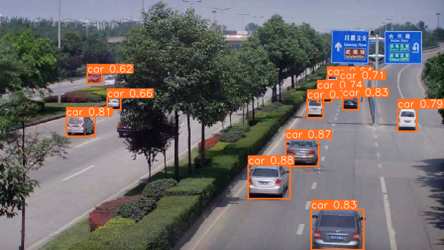}
  \includegraphics[width=3.4cm,height=2.2cm]{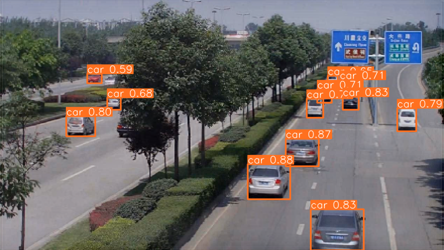}
\leftline{\hspace{2.8cm} (k) GCANet \hspace{1.0cm}  (l) GridDehazeNet \hspace{1.0cm} (m) MSBDN \hspace{1.0cm}  (n) 4KDehazing \hspace{1.0cm}}
\caption{Visual detection results on RTTS (Please zoom in to see the values!).}
\label{fig:detection_visual_RTTS}
\end{figure*}

The results in Table \ref{tab:detection_results_on_RTTS} show that IC-Dehazing achieve the overall best performance. Furthermore, an important and interesting phenomenon is that some dehazing methods lead to the reduction of the object detection performance, while IC-Dehazing does not. Fig. \ref{fig:detection_visual_RTTS} shows the dehazing visual effects of different algorithms and the corresponding object detection results. The first scene contains multiple people, as well as a motorbike on the left. The second scene contains multiple vehicles. According to the visual results, IC-Dehazing does not miss the target (some methods do not detect motorbike successfully). Moreover, IC-Dehazing has relatively high detection confidence.

\begin{table*}
  \centering
  \caption{Quantitative detection results on RTTS.}
  \label{tab:detection_results_on_RTTS}
  \begin{tabular}{c|c|cccccc}
    \hline
    Category                       & Methods        & Person & Bicycle & Car   & Motorbike & Bus   & mAP.5$\uparrow$ \\
    \hline
    \multirow{9}{*}{Non-pair-based} & CANCB          & 0.812 & 0.611 & 0.733 & 0.645 & 0.558 & 0.672 \\
                                   & CEEF           & 0.808 & 0.596 & 0.735 & 0.626 & 0.556 & 0.664 \\
                                   & FME            & 0.806 & \textbf{0.616} & 0.724 & 0.626 & 0.561 & 0.667 \\
                                   & CycleDehaze    & 0.768 & 0.583 & 0.668 & 0.581 & 0.490 & 0.618 \\
                                   & ZID            & 0.791 & 0.551 & 0.724 & 0.584 & 0.532 & 0.636 \\
                                   & SNSPGAN        & 0.811 & 0.614 & 0.726 & \textbf{0.650} & 0.568 & 0.674 \\
                                   & IC-Dehazing    & \textbf{0.815} & 0.608 & \textbf{0.738} & 0.643 & \textbf{0.572} & \textbf{0.675} \\
    \hline
    \multirow{6}{*}{Pair-based}    & DeHamer        & \textbf{0.809} & \textbf{0.615} & \textbf{0.724} & \textbf{0.651} & \textbf{0.559} & \textbf{0.672} \\
                                   & FFA            & 0.804 & 0.599 & 0.717 & 0.630 & 0.547 & 0.659 \\
                                   & GCANet         & 0.795 & 0.601 & 0.697 & 0.626 & 0.535 & 0.651 \\
                                   & GridDehazeNet  & 0.804 & 0.605 & 0.719 & 0.634 & 0.547 & 0.662 \\
                                   & MSBDN          & 0.805 & 0.600 & 0.717 & 0.628 & 0.546 & 0.659 \\
                                   & 4KDehazing     & 0.805 & 0.602 & 0.718 & 0.634 & 0.547 & 0.661 \\
    \hline
  \end{tabular}
\end{table*}

\subsection{Semantic Segmentation Evaluation}
\label{subsec:semantic_segmentation}
Semantic segmentation algorithms based on a single image usually take images containing multiple targets as input and predict the class label corresponding to each pixel position, which is a high-precision semantic understanding task. Since the main purpose here is not to study which semantic segmentation model is more effective, we adopt the classic RefineNet~\cite{lin2017refinenet} as the baseline model.
\begin{figure}
  \centering
  \includegraphics[width=2.8cm,height=2cm]{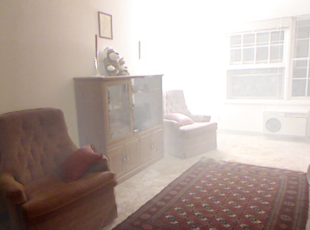}
  \includegraphics[width=2.8cm,height=2cm]{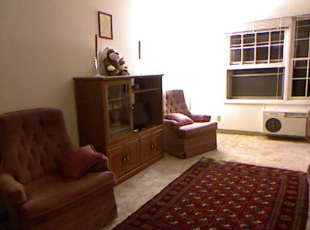}
  \includegraphics[width=2.8cm,height=2cm]{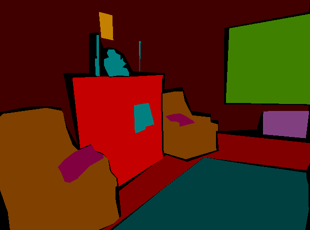}
  \\
  \leftline{\hspace{0.7cm} (a) Hazy \hspace{1.7cm}  (b) Clear \hspace{1.4cm} (c) Label \hspace{1.7cm}}
\caption{Randomly selected example from the NYU-Seg-Haze dataset.}
\label{fig:segmentation_data}
\end{figure}

\begin{figure*}
  \centering
  \includegraphics[width=3.4cm,height=2.4cm]{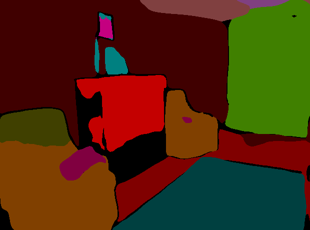}
  \includegraphics[width=3.4cm,height=2.4cm]{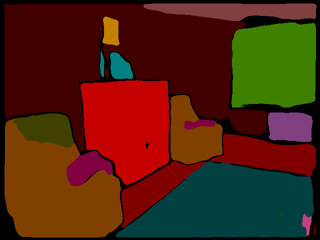}
  \includegraphics[width=3.4cm,height=2.4cm]{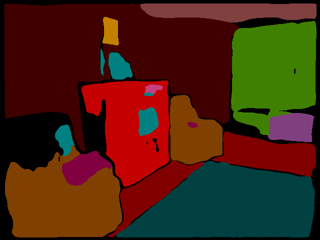}
  \includegraphics[width=3.4cm,height=2.4cm]{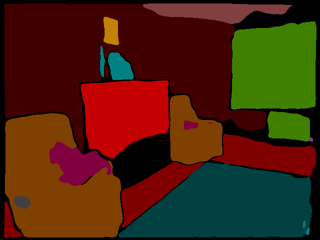}
  \includegraphics[width=3.4cm,height=2.4cm]{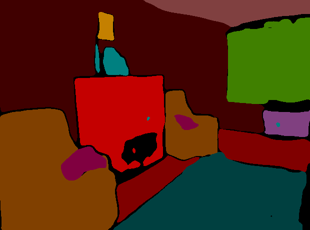}
  \\
  \leftline{\hspace{1.5cm} (a) Hazy \hspace{1.5cm}  (b) CANCB  \hspace{1.7cm}   (c) CEEF  \hspace{2.0cm}  (d) FME  \hspace{1.5cm} (e) CycleDehaze \hspace{1.2cm}  \hspace{1cm}}
  \vspace{0.1cm}
  \includegraphics[width=3.4cm,height=2.4cm]{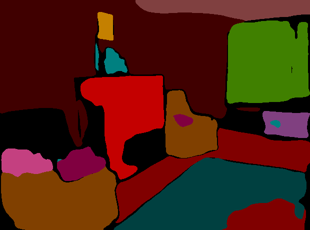}
  \includegraphics[width=3.4cm,height=2.4cm]{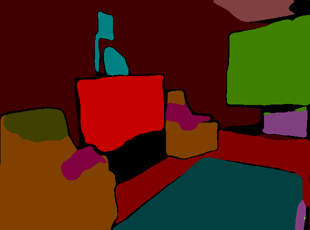}
  \includegraphics[width=3.4cm,height=2.4cm]{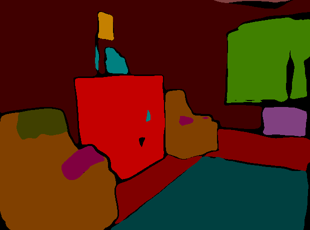}
  \includegraphics[width=3.4cm,height=2.4cm]{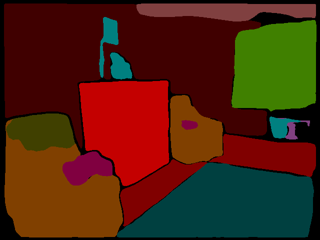}
  \includegraphics[width=3.4cm,height=2.4cm]{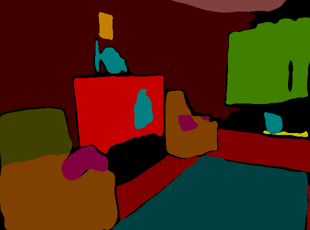}
  \vspace{0.1cm}
  \leftline{\hspace{1.5cm} (f) ZID \hspace{1.7cm}  (g) SNSPGAN \hspace{1.0cm}  (h) IC-Dehazing \hspace{1.1cm}  (i) DeHamer \hspace{1.8cm} (j) FFA \hspace{1.2cm}}
  \\
  \includegraphics[width=3.4cm,height=2.4cm]{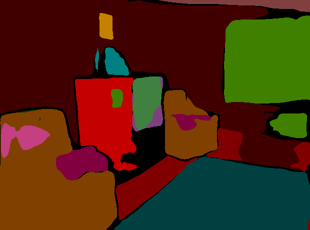}
  \includegraphics[width=3.4cm,height=2.4cm]{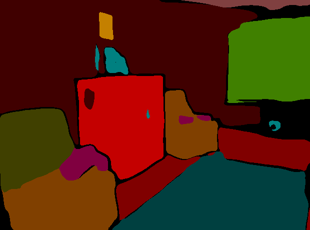}
  \includegraphics[width=3.4cm,height=2.4cm]{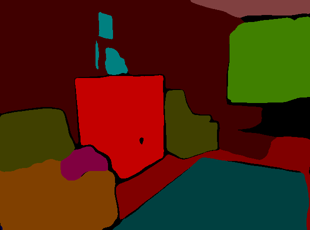}
  \includegraphics[width=3.4cm,height=2.4cm]{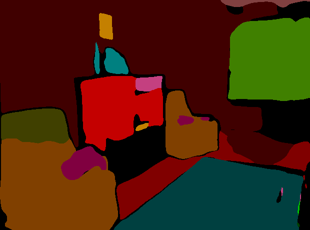}
  \includegraphics[width=3.4cm,height=2.4cm]{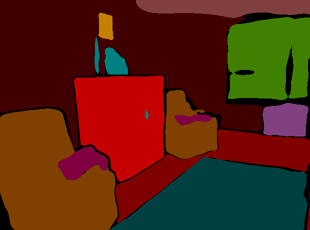}
  \\
  \leftline{\hspace{1.2cm} (k) GCANet \hspace{0.8cm}  (l) GridDehazeNet \hspace{1.0cm}   (m) MSBDN \hspace{1.2cm}  (n) 4KDehazing \hspace{1.5cm} (o) Clear \hspace{1.0cm}}
\caption{Visual segmentation results on NYU-Seg-Haze.}
\label{fig:segmentation_visual_NYU}
\end{figure*}

To perform dehazing and semantic segmentation tasks simultaneously, datasets with accurate depth information and semantic information are required. According to NYU's~\cite{silberman2012indoor} file naming rules, small targets contained in scenes with continuous file name may be duplicated, while scenes with large numerical naming differences have fewer identical targets. For semantic segmentation tasks, the target distributions for training and testing should be as consistent as possible. The scene of ITS~\cite{li2018benchmarking} comes from NYU, but the validation set of ITS comes from 50 consecutive images of NYU. Thus, it is not suitable as a dataset for the dehazing $\&$ semantic segmentation joint task. Therefore, D-Hazy~\cite{ancuti2016d} and NYU~\cite{silberman2012indoor} are chosen as evaluation datasets for dehazing and segmentation, which are called NYU-Seg-Haze. NYU can provide 40 classes of objects, which is a very complex dataset containing many small objects. For the evaluation of segmentation results, mean pixel accuracy (mPA) and mean intersection over union (mIoU) are selected as quantitative metrics.

Semantic segmentation experimental results on Table \ref{tab:segmentation_results_on_NYU} show that the negative impact of haze on pixel-level understanding of images is obvious. The mPA and mIoU obtained when the clear image is used as input to the segmentation network are 0.571 and 0.445, respectively. However, mPA and mIoU are only 0.421 and 0.303 when hazy images are input to the algorithm. Although the PSNR and SSIM values obtained by IC-Dehazing on the NYU-Seg-Haze dataset do not exceed the best supervised algorithm, it achieves the second highest segmentation quantitative evaluation values of 0.463 and 0.353, respectively. The possible reason is that the dehazing module of IC-Dehazing is composed of a window-based attention mechanism, which is more natural for the restoration of local objects. Fig. \ref{fig:segmentation_visual_NYU} shows the segmentation results of the scene corresponding to Fig. \ref{fig:segmentation_data}. In hazy conditions, there is a large area of wrong segmentation of sofas and cabinets. Misrecognition is the least on the clear image, which are the closest to the ground-truth semantic label. The segmentation results corresponding to various dehazing algorithms are misidentified at the location of the wall (the top of the picture), but IC-Dehazing does not. Overall, IC-Dehazing achieves impressive object segmentation performance. 

\begin{table}
  \centering
  \caption{Quantitative segmentation results on NYU-Seg-Haze.}
  \label{tab:segmentation_results_on_NYU}
  \begin{tabular}{c|c|cc}
    \hline
    Category                       & Methods        & mPA$\uparrow$ & mIoU$\uparrow$ \\
    \hline
    \multirow{9}{*}{Non-pair-based} & Clear          & 0.571 & 0.445 \\
                                   & Hazy           & 0.421 & 0.303 \\
    \hline
                                   & CANCB          & 0.441 & 0.334 \\
                                   & CEEF           & 0.436 & 0.322 \\
                                   & FME            & 0.438 & 0.320 \\
                                   & CycleDehaze    & 0.429 & 0.310 \\
                                   & ZID            & 0.347 & 0.259 \\
                                   & SNSPGAN        & 0.402 & 0.289 \\
                                   & IC-Dehazing    & \textbf{0.463} & \textbf{0.353} \\
    \hline
    \multirow{6}{*}{Pair-based}    & DeHamer        & \textbf{0.467} & \textbf{0.357} \\
                                   & FFA            & 0.410 & 0.307 \\
                                   & GCANet         & 0.395 & 0.282 \\
                                   & GridDehazeNet  & 0.426 & 0.312 \\
                                   & MSBDN          & 0.408 & 0.306 \\
                                   & 4KDehazing     & 0.404 & 0.290 \\
    \hline
  \end{tabular}
\end{table}

\subsection{Ablation Study}
This section will analyze the role of various components included in IC-Dehazing. As introduced in this paper, IC-Dehazing mainly includes three components. First, a dehazing network which is built based on Transformer. Second, a CI Module for obtaining different lighting outputs. Third, a prior loss that can assist unsupervised model training. For simplicity, the above three items are denoted as Former, CI and Prior, respectively. Therefore, the purpose of the ablation experiment is to prove that Former and Prior can promote the overall performance of dehazing. It is worth noting that the role of CI is to obtain different illuminations, which is not intended to improve quantitative results. The training process of IC-Dehazing is inspired by CycleGAN,~\cite{zhu2017unpaired} denoted as Res, which represents CycleGAN built by the residual generator. Therefore, the overall ablation experiment consists of 4 parts:
\begin{itemize}
  \item (A) Res: Architecture that uses CycleGAN entirely.
  \item (B) Former: Replace the generator of Res with the Transformer architecture, which is equivalent to HSM.
  \item (C) Former + CI: Add CI module to the Former.
  \item (D) Former + CI + Prior: IC-Dehazing with complete configuration.
\end{itemize}

\begin{table*}
  \centering
  \caption{Ablation study on prior loss and illumination controllable module.}
  \label{tab:ablation_study}
  \begin{tabular}{ccccccc}
    \hline
    \multirow{2}{*}{Config} & \multicolumn{2}{c}{ITS} & \multicolumn{2}{c}{OTS} & \multicolumn{2}{c}{4K} \\
    \cline{2-7}
    & SSIM$\uparrow$ & PSNR$\uparrow$ & SSIM$\uparrow$ & PSNR$\uparrow$ & SSIM$\uparrow$ & PSNR$\uparrow$ \\
    \hline
    A: Res & 0.793 &  19.332 & 0.885 & 23.875 & 0.856 & 22.112 \\
    B: Former & 0.880 & 21.545 & 0.920 & 23.774 & 0.921 & 22.546 \\
    C: Former + CI & 0.873  & 20.665 & 0.918 & 23.717 & 0.917 & 22.364 \\
    D: Former + CI + Prior & \textbf{0.894} & \textbf{22.558} & \textbf{0.929} & \textbf{24.562} & \textbf{0.936} & \textbf{23.355} \\
    \hline
  \end{tabular}
\end{table*}

Ablation experiments use indoor ITS and outdoor OTS/4K datasets, and the hyperparameters such as learning rate and batch size are consistent for all experiments. The quantitative results in Table \ref{tab:ablation_study} suggest three important conclusions. First, compared A with B, we can conclude that using a Transformer-based architecture performs better than a conventional convolution-based architecture. Second, compared to configuration B, configuration C adds CI, which has a slight negative impact on the model. This result is reasonable, since the purpose of CI is not to improve the performance of the model, but to obtain the output of different illuminations, so it may increase the difficulty of model training. Third, compared to B and C, D is able to further improve the performance of the dehazing network with CI and Former, which proves that Prior is effective.

Further, we compare the difference between computing prior loss in feature space and pixel space. The results are shown in Table \ref{tab:ablation_study_pixel_feature}. The quality of the dehazed image is determined by the reconstruction loss, adversarial loss and prior loss. The reconstruction loss is at the pixel level, and the adversarial loss and the prior loss are at the feature level. From the experimental results, it can be seen that computing the prior loss in the feature space is a better strategy than pixel space. The reason is that prior knowledge can reflect the overall feature level statistical laws of a large number of images, rather than precisely for each pixel. Therefore, it is a reasonable strategy to use feature-level prior loss and adversarial loss with pixel-level reconstruction loss.
\begin{table}
  \scriptsize
  \centering
  \caption{Ablation study that computes prior loss in pixel space and feature space.}
  \label{tab:ablation_study_pixel_feature}
  \begin{tabular}{ccccccc}
    \hline
    \multirow{2}{*}{Config} & \multicolumn{2}{c}{ITS} & \multicolumn{2}{c}{OTS} & \multicolumn{2}{c}{4K} \\
    \cline{2-7}
    & SSIM$\uparrow$ & PSNR$\uparrow$ & SSIM$\uparrow$ & PSNR$\uparrow$ & SSIM$\uparrow$ & PSNR$\uparrow$ \\
    \hline
    Pixel & 0.881 & \textbf{22.571} & 0.915 & 23.050 & 0.914 & 22.789 \\
    Feature & \textbf{0.894} & 22.558 & \textbf{0.929} & \textbf{24.562} & \textbf{0.936} & \textbf{23.355} \\
    \hline
  \end{tabular}
\end{table}

\subsection{Analysis of Illumination Controllable Property}
In this section, four aspects of lighting controllable property will be studied. First, the visual dehazing effects obtained by different $\alpha$ are shown. Second, a quantitative evaluation of different $\alpha$ is performed to demonstrate that the structural information of the image is not destroyed when $\alpha$ is changed. Third, whether the Transformer network has learned illumination map $L_{ir}$ will be verified. Fourth, using different weights $\alpha$ for $L_{ir}$, the pixel distribution relationship between the predicted image and the clear image will be studied.

\subsubsection{Visual Results of Different Illumination Parameters}
Fig. \ref{fig:diff_alpha_ITS}, Fig. \ref{fig:diff_alpha_OTS} and Fig. \ref{fig:diff_alpha_RTTS} show the visual results corresponding to different $\alpha$. It can be seen that as $\alpha$ increases, the lighting of the scene changes and a series of acceptable dehazed images are produced. This proves that IC-Dehazing, which provides multiple outputs, is effective for ill-posedness dehazing problem.

\begin{figure*}
  \centering
  \includegraphics[width=2.4cm,height=1.7cm]{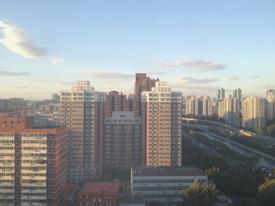}
  \includegraphics[width=2.4cm,height=1.7cm]{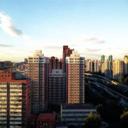}
  \includegraphics[width=2.4cm,height=1.7cm]{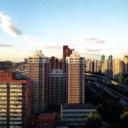}
  \includegraphics[width=2.4cm,height=1.7cm]{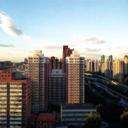}
  \includegraphics[width=2.4cm,height=1.7cm]{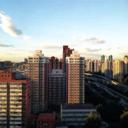}
  \includegraphics[width=2.4cm,height=1.7cm]{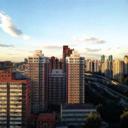}
  \includegraphics[width=2.4cm,height=1.7cm]{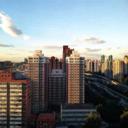}
  \vspace{0.1cm}
  \leftline{\hspace{0.8cm} (a) Hazy \hspace{1.2cm}  (b) 0.42 \hspace{1.0cm}   (c) 0.44  \hspace{1.2cm}  (d) 0.45  \hspace{1.2cm} (e) 0.46  \hspace{1.2cm}  (f) 0.47  \hspace{1.0cm} (g) 0.48}
  \\
  \includegraphics[width=2.4cm,height=1.7cm]{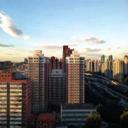}
  \includegraphics[width=2.4cm,height=1.7cm]{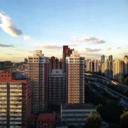}
  \includegraphics[width=2.4cm,height=1.7cm]{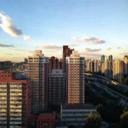}
  \includegraphics[width=2.4cm,height=1.7cm]{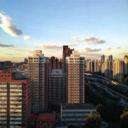}
  \includegraphics[width=2.4cm,height=1.7cm]{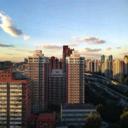}
  \includegraphics[width=2.4cm,height=1.7cm]{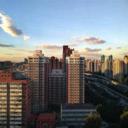}
  \includegraphics[width=2.4cm,height=1.7cm]{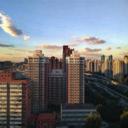}
  \\
  \leftline{\hspace{1.0cm}  (h) 0.49 \hspace{1.0cm}   (i) 0.50 \hspace{1.2cm}  (j) 0.51 \hspace{1.2cm} (k) 0.52 \hspace{1.2cm}  (l) 0.53  \hspace{1.2cm} (m) 0.54 \hspace{1.0cm} (n) 0.56}

\rule[0.4\baselineskip]{\textwidth}{0.5pt}

  \includegraphics[width=2.4cm,height=1.7cm]{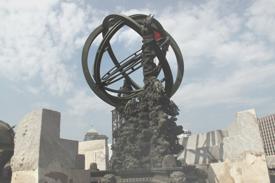}
  \includegraphics[width=2.4cm,height=1.7cm]{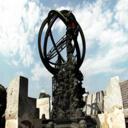}
  \includegraphics[width=2.4cm,height=1.7cm]{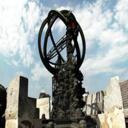}
  \includegraphics[width=2.4cm,height=1.7cm]{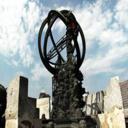}
  \includegraphics[width=2.4cm,height=1.7cm]{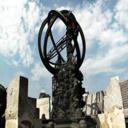}
  \includegraphics[width=2.4cm,height=1.7cm]{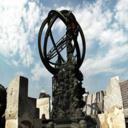}
  \includegraphics[width=2.4cm,height=1.7cm]{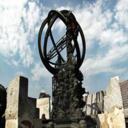}
  \vspace{0.1cm}
  \leftline{\hspace{0.8cm} (a) Hazy \hspace{1.2cm}  (b) 0.42 \hspace{1.0cm}   (c) 0.44  \hspace{1.2cm}  (d) 0.45  \hspace{1.2cm} (e) 0.46  \hspace{1.2cm}  (f) 0.47  \hspace{1.0cm} (g) 0.48}
  \\
  \includegraphics[width=2.4cm,height=1.7cm]{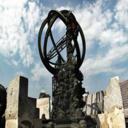}
  \includegraphics[width=2.4cm,height=1.7cm]{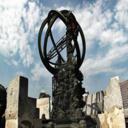}
  \includegraphics[width=2.4cm,height=1.7cm]{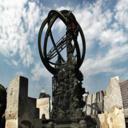}
  \includegraphics[width=2.4cm,height=1.7cm]{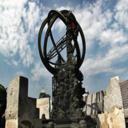}
  \includegraphics[width=2.4cm,height=1.7cm]{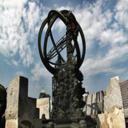}
  \includegraphics[width=2.4cm,height=1.7cm]{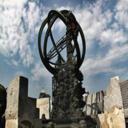}
  \includegraphics[width=2.4cm,height=1.7cm]{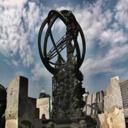}
  \\
  \leftline{\hspace{1.0cm}  (h) 0.49 \hspace{1.0cm}   (i) 0.50 \hspace{1.2cm}  (j) 0.51 \hspace{1.2cm} (k) 0.52 \hspace{1.2cm}  (l) 0.53  \hspace{1.2cm} (m) 0.54 \hspace{1.0cm} (n) 0.56}

\rule[0.4\baselineskip]{\textwidth}{0.5pt}

  \includegraphics[width=2.4cm,height=1.7cm]{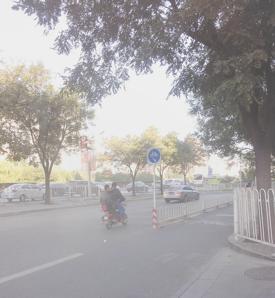}
  \includegraphics[width=2.4cm,height=1.7cm]{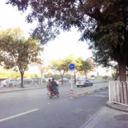}
  \includegraphics[width=2.4cm,height=1.7cm]{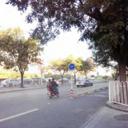}
  \includegraphics[width=2.4cm,height=1.7cm]{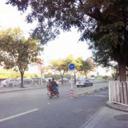}
  \includegraphics[width=2.4cm,height=1.7cm]{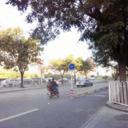}
  \includegraphics[width=2.4cm,height=1.7cm]{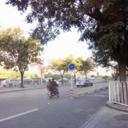}
  \includegraphics[width=2.4cm,height=1.7cm]{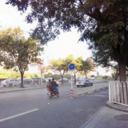}
  \vspace{0.1cm}
  \leftline{\hspace{0.8cm} (a) Hazy \hspace{1.2cm}  (b) 0.42 \hspace{1.0cm}   (c) 0.44  \hspace{1.2cm}  (d) 0.45  \hspace{1.2cm} (e) 0.46  \hspace{1.2cm}  (f) 0.47  \hspace{1.0cm} (g) 0.48}
  \\
  \includegraphics[width=2.4cm,height=1.7cm]{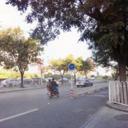}
  \includegraphics[width=2.4cm,height=1.7cm]{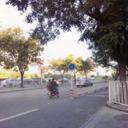}
  \includegraphics[width=2.4cm,height=1.7cm]{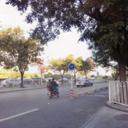}
  \includegraphics[width=2.4cm,height=1.7cm]{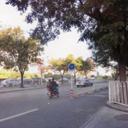}
  \includegraphics[width=2.4cm,height=1.7cm]{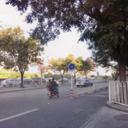}
  \includegraphics[width=2.4cm,height=1.7cm]{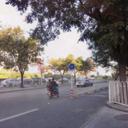}
  \includegraphics[width=2.4cm,height=1.7cm]{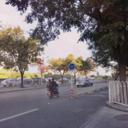}
  \\
  \leftline{\hspace{1.0cm}  (h) 0.49 \hspace{1.0cm}   (i) 0.50 \hspace{1.2cm}  (j) 0.51 \hspace{1.2cm} (k) 0.52 \hspace{1.2cm}  (l) 0.53  \hspace{1.2cm} (m) 0.54 \hspace{1.0cm} (n) 0.56}

\rule[0.4\baselineskip]{\textwidth}{0.5pt}

  \includegraphics[width=2.4cm,height=1.7cm]{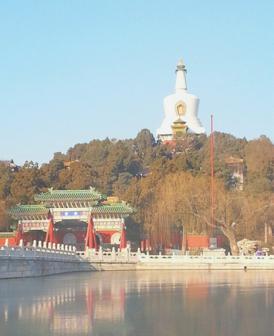}
  \includegraphics[width=2.4cm,height=1.7cm]{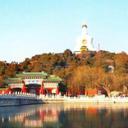}
  \includegraphics[width=2.4cm,height=1.7cm]{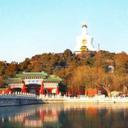}
  \includegraphics[width=2.4cm,height=1.7cm]{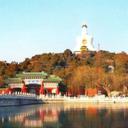}
  \includegraphics[width=2.4cm,height=1.7cm]{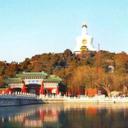}
  \includegraphics[width=2.4cm,height=1.7cm]{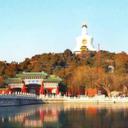}
  \includegraphics[width=2.4cm,height=1.7cm]{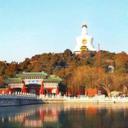}
  \vspace{0.1cm}
  \leftline{\hspace{0.8cm} (a) Hazy \hspace{1.2cm}  (b) 0.42 \hspace{1.0cm}   (c) 0.44  \hspace{1.2cm}  (d) 0.45  \hspace{1.2cm} (e) 0.46  \hspace{1.2cm}  (f) 0.47  \hspace{1.0cm} (g) 0.48}
  \\
  \includegraphics[width=2.4cm,height=1.7cm]{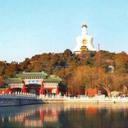}
  \includegraphics[width=2.4cm,height=1.7cm]{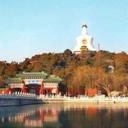}
  \includegraphics[width=2.4cm,height=1.7cm]{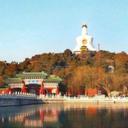}
  \includegraphics[width=2.4cm,height=1.7cm]{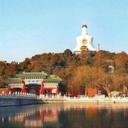}
  \includegraphics[width=2.4cm,height=1.7cm]{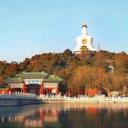}
  \includegraphics[width=2.4cm,height=1.7cm]{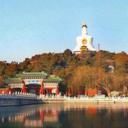}
  \includegraphics[width=2.4cm,height=1.7cm]{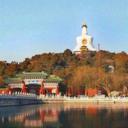}
  \\
  \leftline{\hspace{1.0cm}  (h) 0.49 \hspace{1.0cm}   (i) 0.50 \hspace{1.2cm}  (j) 0.51 \hspace{1.2cm} (k) 0.52 \hspace{1.2cm}  (l) 0.53  \hspace{1.2cm} (m) 0.54 \hspace{1.0cm} (n) 0.56}
  \caption{Visual dehazing results with different controllable parameters. From top to bottom in the figure, there are four different scenes divided by horizontal lines and marked as A, B, C and D respectively. It can be seen that when selecting different controllable parameters, there are differences in the illumination of the corresponding images.}
  \label{fig:diff_alpha_OTS_supp}
\end{figure*}

More results on testset~\cite{li2018benchmarking} are shown in Fig. \ref{fig:diff_alpha_OTS_supp}. There are illumination differences in the visual effects of sky, lake and buildings in the same scene of different images.

\subsubsection{Quantitative Evaluation of Different Illumination Parameters}
Table \ref{tab:quantitative_differ_alpha} shows the quantitative evaluation results obtained on different datasets for different $\alpha$. On one hand, it shows that dehazing works best when $\alpha$ is equal to 0.5. This conclusion is expected since $\alpha$ is set to 0.5 during network optimization and the ground-truth image is uniquely determined. On the other hand, Table \ref{tab:quantitative_differ_alpha} also shows that when $\alpha$ varies, that is, when dehazed images with different illuminations are generated, the dehazed images corresponding to each $\alpha$ and the ground-truth still maintain a high degree of structural similarity. This result proves that the contents of the dehazed images are not destroyed when IC-Dehazing changes the illumination value.
\begin{table}
  \scriptsize
  \centering
  \caption{Quantitative dehazing results of different $\alpha$.}
  \label{tab:quantitative_differ_alpha}
  \begin{tabular}{ccccccc}
    \hline
    \multirow{2}{*}{Parameters} & \multicolumn{2}{c}{ITS} & \multicolumn{2}{c}{OTS} & \multicolumn{2}{c}{4K} \\
    \cline{2-7}
    & SSIM$\uparrow$ & PSNR$\uparrow$ & SSIM$\uparrow$ & PSNR$\uparrow$ & SSIM$\uparrow$ & PSNR$\uparrow$ \\
    \hline
     $\alpha=0.42$ & 0.850 & 18.559 & 0.906 & 21.537 & 0.909 & 19.814 \\
     $\alpha=0.44$ & 0.868 & 19.900 & 0.915 & 22.449 & 0.919 & 21.028 \\
     $\alpha=0.45$ & 0.875 & 20.580 & 0.918 & 22.886 & 0.924 & 21.637 \\
     $\alpha=0.46$ & 0.882 & 21.230 & 0.922 & 23.358 & 0.928 & 22.211 \\
     $\alpha=0.47$ & 0.888 & 21.809 & 0.925 & 23.773 & 0.931 & 22.707 \\
     $\alpha=0.48$ & 0.891 & 22.263 & 0.928 & 24.126 & 0.934 & 23.082 \\
     $\alpha=0.49$ & \textbf{0.894} & 22.532 & 0.929 & 24.385 & 0.935 & 23.304 \\
     $\alpha=0.50$ & \textbf{0.894} & \textbf{22.558} & 0.929 & \textbf{24.562} & \textbf{0.936} & \textbf{23.355}\\
     $\alpha=0.51$ & 0.892 & 22.311 & \textbf{0.930} & 24.481 & 0.935 & 23.227\\
     $\alpha=0.52$ & 0.888 & 21.810 & 0.928 & 24.262 & 0.934 & 22.931 \\
     $\alpha=0.53$ & 0.881 & 21.121 & 0.927 & 23.846 & 0.931 & 22.504 \\
     $\alpha=0.54$ & 0.872 & 20.317 & 0.924 & 23.240 & 0.926 & 21.991 \\
     $\alpha=0.56$ & 0.844 & 18.557 & 0.912 & 21.576 & 0.914 & 20.862 \\
    \hline
  \end{tabular}
\end{table}
\subsubsection{Visualization of Illumination Map}
\begin{figure*}
  \centering
  \includegraphics[width=2.8cm,height=2cm]{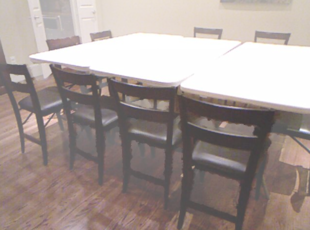}
  \includegraphics[width=2.8cm,height=2cm]{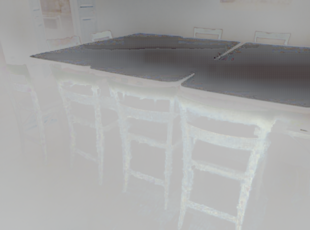}
  \includegraphics[width=2.8cm,height=2cm]{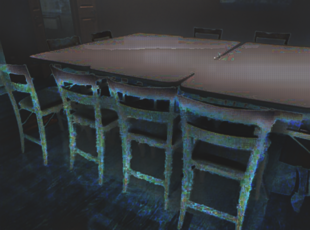}
  \includegraphics[width=2.8cm,height=2cm]{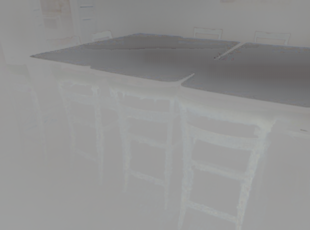}
  \includegraphics[width=2.8cm,height=2cm]{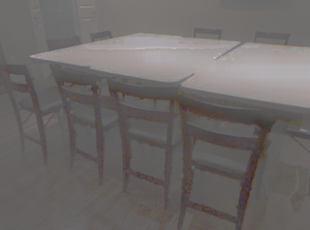}
  \includegraphics[width=2.8cm,height=2cm]{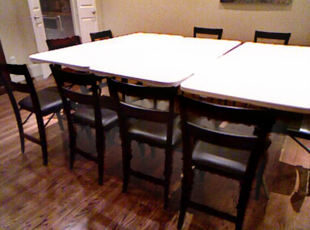}
\\
\vspace{0.1cm}
  \includegraphics[width=2.8cm,height=2.9cm]{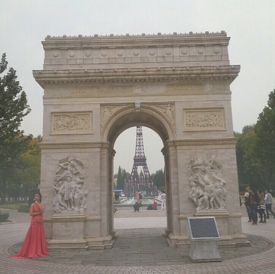}
  \includegraphics[width=2.8cm,height=2.9cm]{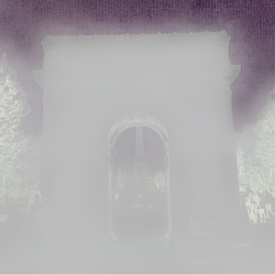}
  \includegraphics[width=2.8cm,height=2.9cm]{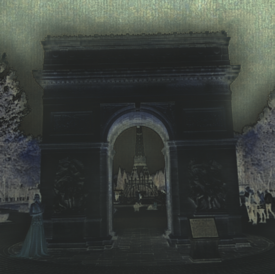}
  \includegraphics[width=2.8cm,height=2.9cm]{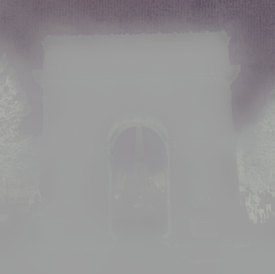}
  \includegraphics[width=2.8cm,height=2.9cm]{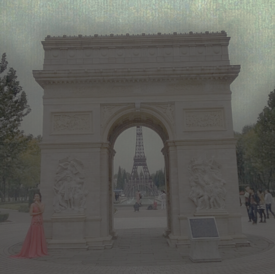}
  \includegraphics[width=2.8cm,height=2.9cm]{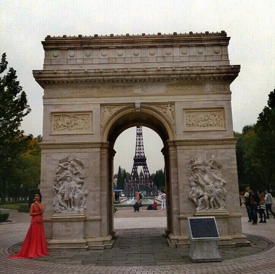}
\\
\vspace{0.1cm}
  \includegraphics[width=2.8cm,height=1.6cm]{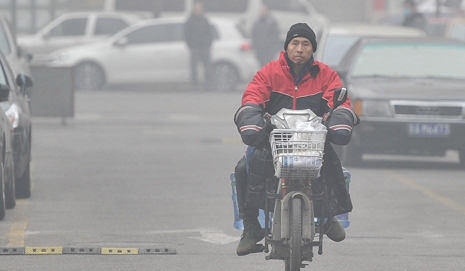}
  \includegraphics[width=2.8cm,height=1.6cm]{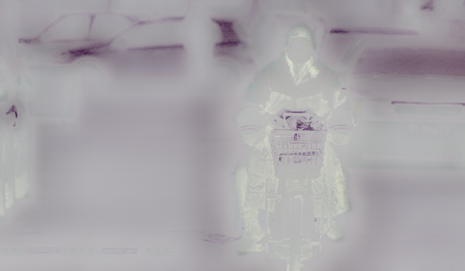}
  \includegraphics[width=2.8cm,height=1.6cm]{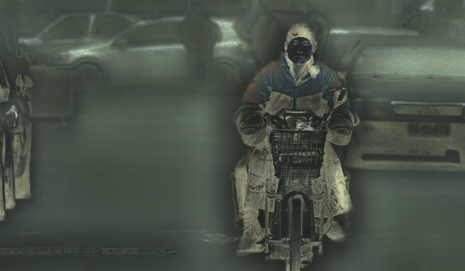}
  \includegraphics[width=2.8cm,height=1.6cm]{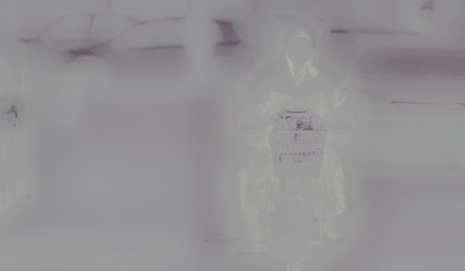}
  \includegraphics[width=2.8cm,height=1.6cm]{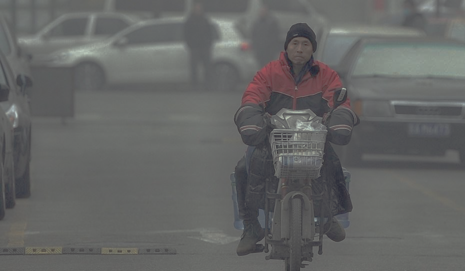}
  \includegraphics[width=2.8cm,height=1.6cm]{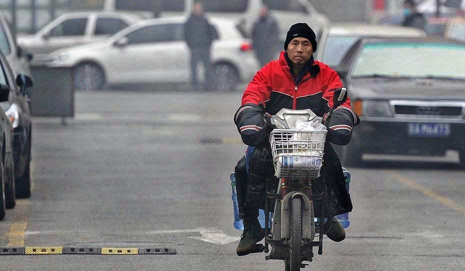}
  \leftline{\hspace{1.0cm} (a) Hazy \hspace{1.5cm}  (b) $o_{d}(x)$ \hspace{1.5cm}  (c) $L_{ir}$ \hspace{1.1cm}  (d) $\alpha \times o_{d}(x)$  \hspace{0.25cm} (e) $(1 - \alpha) \times o_ {r}(x)$ \hspace{0.35cm} (f) Prediction}
\caption{Illumination maps and intermediate outputs.}
\label{fig:CI_map}
\end{figure*}
As introduced in this paper, $o_{d}(x)$ and $L_{ir} = o_{r}(x) / x$ can be obtained when the model performs forward computing. Since $o_{d}(x)$ and $L_{ir}$ are not normalized, their numerical interval may be arbitrary. To visualize $o_{d}(x)$, $L_{ir}$ and $o_{r}(x)$, max-min normalization is used to obtain image formats with values ranging from 0 to 255. Further, the weighted $o_{d}(x)$ and $o_{r}(x)$ are also visualized, that is $\alpha \times o_{d}(x)$ and $(1 - \alpha) \times o_ {r}(x)$. For an image $u$, the max-min normalization can be expressed as
\begin{equation}
  u_{norm}(i,j) = \frac{u(i, j) - min(u)}{max(u) - min(u)} * 255, 
\end{equation}
where $i$ and $j$ denotes pixel location. The $L_{ir}$ shown in Fig. \ref{fig:CI_map} can express the brightness difference information of different objects in the image (where $\alpha$ is chosen as $0.5$). Moreover, $(1 - \alpha) \times o_ {r}(x)$ represents the overall content of the image and the outline details of the object. By combining $\alpha \times o_{d}(x)$ and $(1 - \alpha) \times o_{r}(x)$, the haze in the final predicted image is removed.

\subsubsection{Pixel Histogram}
\begin{figure*}
  \centering
  \includegraphics[width=4.4cm,height=3.3cm]{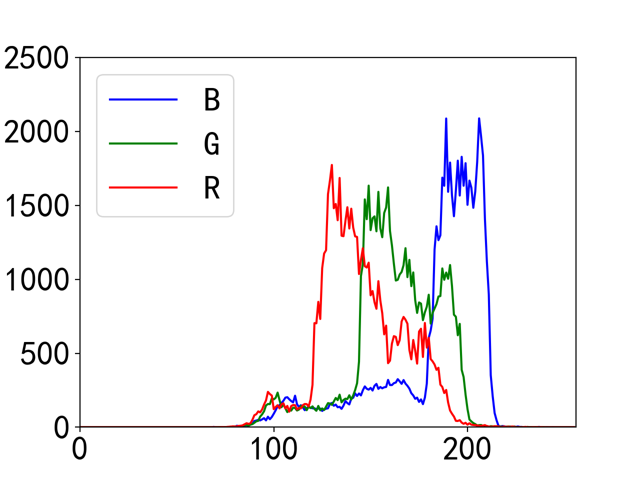}
  \includegraphics[width=4.4cm,height=3.3cm]{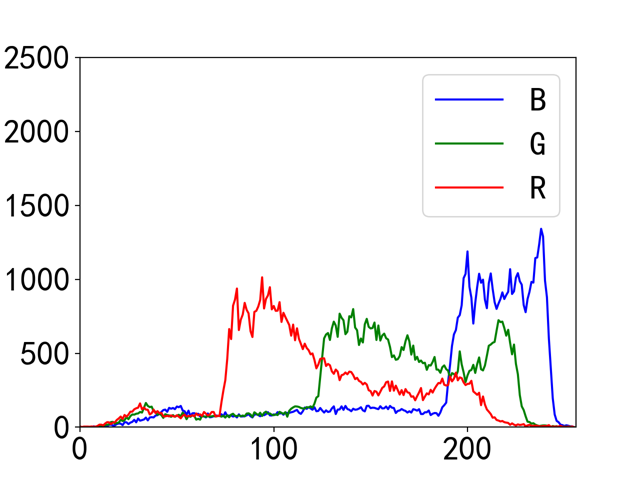}
  \includegraphics[width=4.4cm,height=3.3cm]{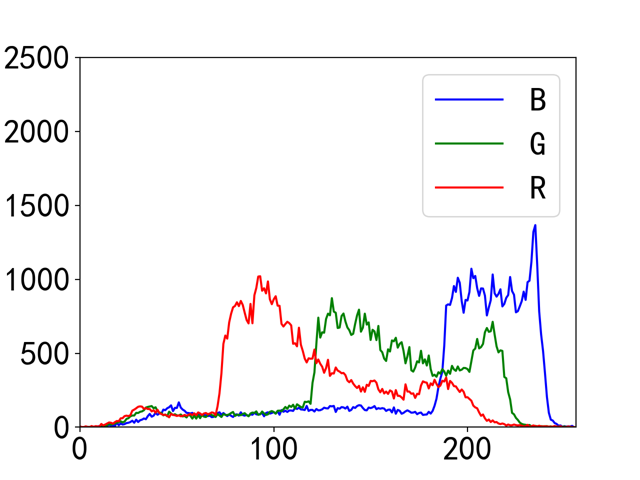}
  \includegraphics[width=4.4cm,height=3.3cm]{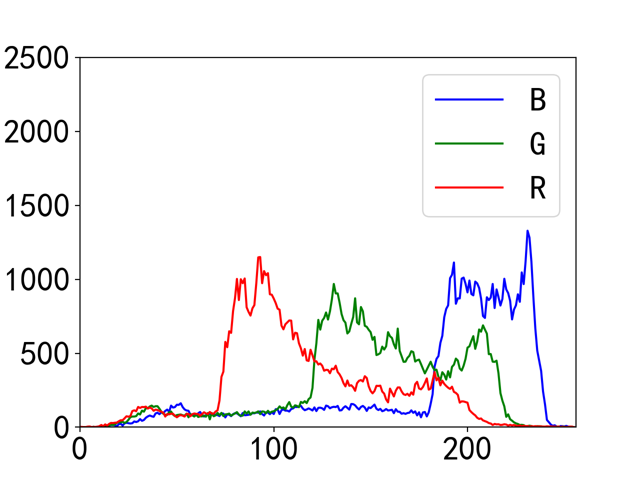}
  \vspace{0.1cm}
  \leftline{\hspace{1.5cm} (a) Hazy \hspace{2.5cm}  (b)  $\alpha=0.42$  \hspace{2.7cm}  (c) $\alpha=0.44$ \hspace{2.5cm} (d) $\alpha=0.45$ \hspace{0.7cm}}
  \\
  \includegraphics[width=4.4cm,height=3.3cm]{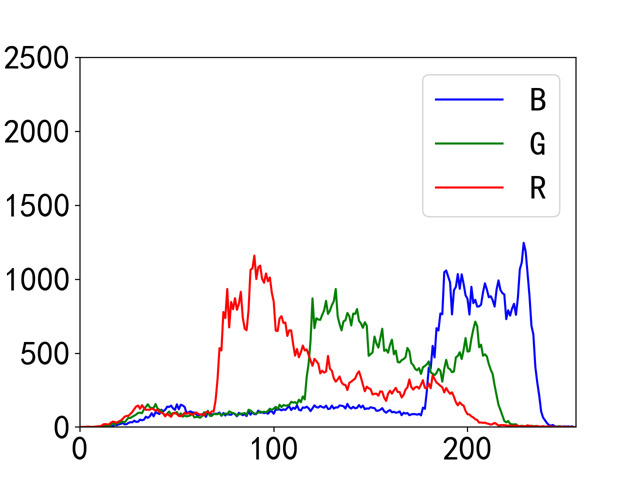}
  \includegraphics[width=4.4cm,height=3.3cm]{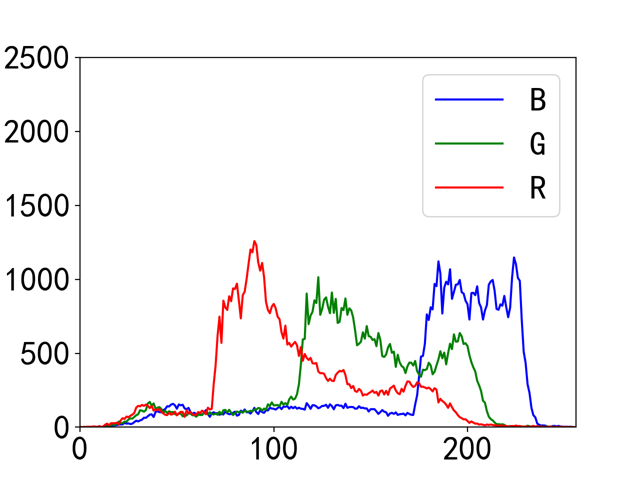}
  \includegraphics[width=4.4cm,height=3.3cm]{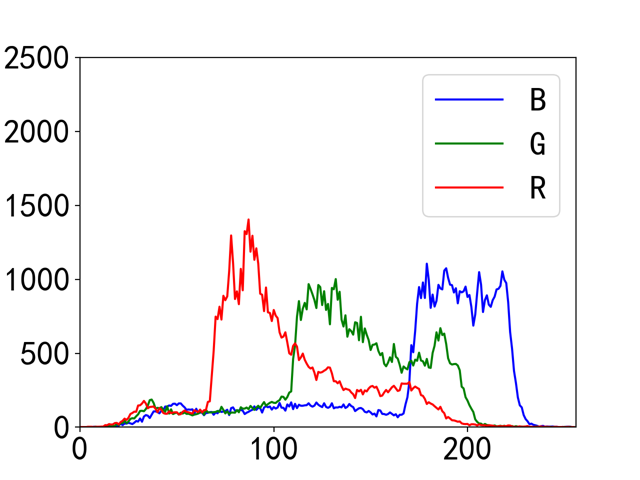}
  \includegraphics[width=4.4cm,height=3.3cm]{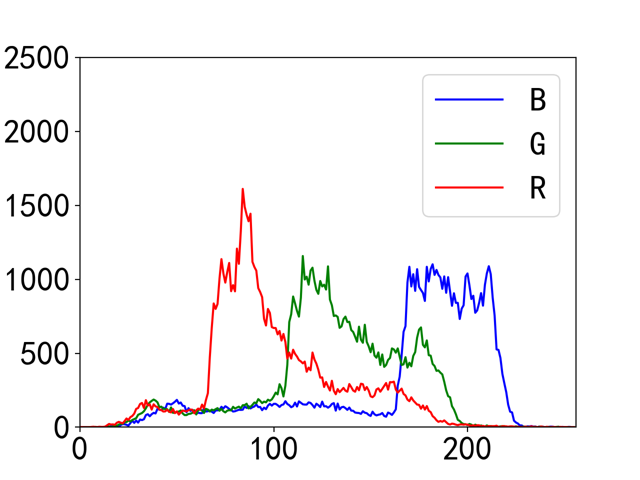}
  \\
  \leftline{\hspace{1.5cm} (e) $\alpha=0.46$  \hspace{2.3cm}  (f)  $\alpha=0.48$  \hspace{2.5cm}  (g) $\alpha=0.50$ \hspace{2.3cm} (h) $\alpha=0.52$ \hspace{0.7cm}}
  \includegraphics[width=4.4cm,height=3.3cm]{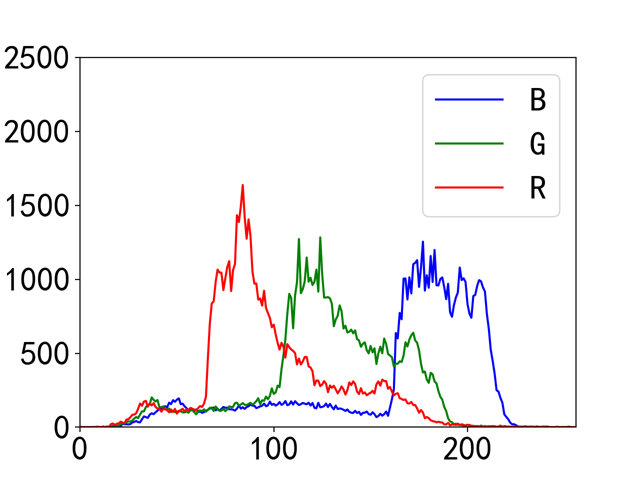}
  \includegraphics[width=4.4cm,height=3.3cm]{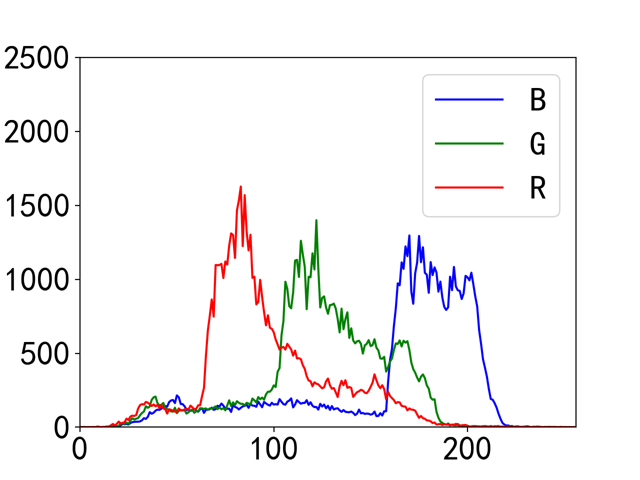}
  \includegraphics[width=4.4cm,height=3.3cm]{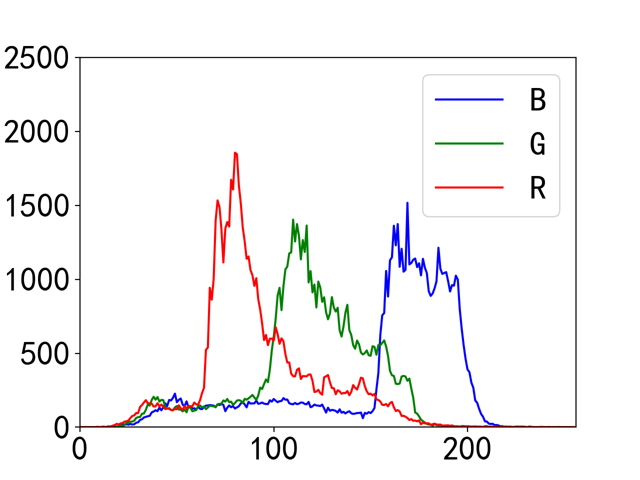}
  \includegraphics[width=4.4cm,height=3.3cm]{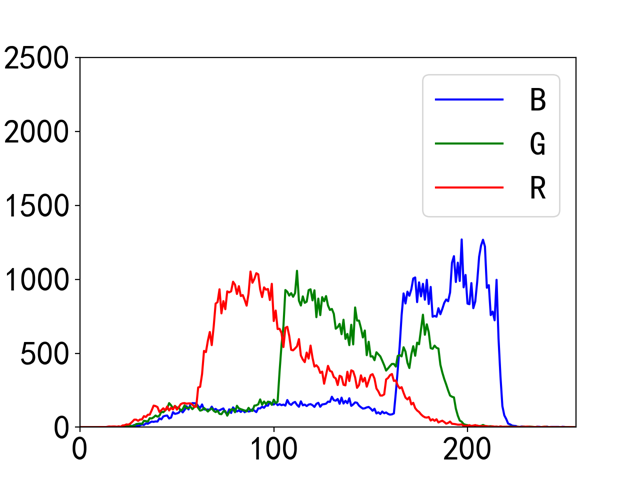}
  \\
  \leftline{\hspace{1.5cm} (i) $\alpha=0.53$  \hspace{2.3cm}  (j)  $\alpha=0.54$  \hspace{2.5cm}  (k) $\alpha=0.56$ \hspace{2.3cm} (l) Clear \hspace{0.7cm}}
\caption{Pixel histograms corresponding to different $\alpha$ in the OTS dataset.}
\label{fig:OTS_diff_alpha_hist}
\end{figure*}

As shown in Fig. \ref{fig:diff_alpha_OTS}, there are obvious visual effect differences among the images when different $\alpha$ are selected. To quantitatively investigate the source of this visual difference, pixel histograms corresponding to different predicted images are plotted. The results shown in Fig. \ref{fig:OTS_diff_alpha_hist} indicate that the number and distribution of pixels of the Red (R), Green (G) and Blue (B) components corresponding to different $\alpha$ are different. When $\alpha=0.44$, the relative distribution trends of the three color components are closest to the true labels (Clear). When $\alpha=0.56$, the proportion of the red component is significantly higher. Therefore, the pixel distribution changes from R, G and B reflect why the visual effects corresponding to different $\alpha$ are different.

\section{Conclusion}
This paper proposes an illumination controllable unsupervised dehazing model called IC-Dehazing. To obtain high-quality dehazing results and control the illumination conditions of images, IC-Dehazing designs three effective strategies include the encoder-decoder dehazing network with the Transformer architecture, the illumination control module and the prior loss. To verify the performance of the proposed model, quantitative and qualitative experiments are conducted on dehazing, object detection and semantic segmentation. The results show that the proposed IC-Dehazing can obtain the state-of-the-art unsupervised dehazing effect.

\section*{Acknowledgment}
We thank the Big Data Computing Center of Southeast University for providing the facility support on the numerical calculations in this paper. This work was supported in part by National Key R\&D Program of China under Grant 2022YFB3104301; the grant of the National Science Foundation of China under Grant 62172090; Start-up Research Fund of Southeast University under Grant RF1028623097; CAAI-Huawei MindSpore Open Fund. All correspondence should be directed to Xiaofeng Cong.
\bibliography{refs}
\end{document}